\documentclass{article}

\usepackage[preprint]{neurips_2025}

\PassOptionsToPackage{authoryear}{natbib}
\usepackage{natbib}
\usepackage[utf8]{inputenc} %
\usepackage[T1]{fontenc}    %
\usepackage{hyperref}       %
\usepackage{url}            %
\usepackage{booktabs}       %
\usepackage{amsfonts}       %
\usepackage{nicefrac}       %
\usepackage{microtype}      %
\usepackage{xcolor}         %
\usepackage{graphicx}       %
\usepackage{multirow}       %
\usepackage{tabularx}
\usepackage{wrapfig}
\usepackage[normalem]{ulem}
\useunder{\uline}{\ul}{}

\title{DrafterBench: Benchmarking Large Language Models for Tasks Automation in Civil Engineering}

\author{%
  Yinsheng Li\\
  Department of Civil Engineering\\
  McGill University\\
  \texttt{yinsheng.li@mail.mcgill.ca} \\
  \AND
  Zhen Dong\\
  UC Santa Barbara and NVIDIA\\
  \texttt{zhendong@berkeley.edu} \\
  \AND
  Yi Shao\\
  Department of Civil Engineering\\
  McGill University\\
  \texttt{yi.shao2@mcgill.ca} \\
}

\begin{document}

\maketitle

\begin{abstract}
  Large Language Model (LLM) agents have shown great potential for solving real-world problems and promise to be a solution for tasks automation in industry. However, more benchmarks are needed to systematically evaluate automation agents from an industrial perspective, for example, in Civil Engineering. Therefore, we propose DrafterBench for the comprehensive evaluation of LLM agents in the context of technical drawing revision, a representation task in civil engineering. DrafterBench contains twelve types of tasks summarized from real-world drawing files, with 46 customized functions/tools and 1920 tasks in total. DrafterBench is an open-source benchmark to rigorously test AI agents' proficiency in interpreting intricate and long-context instructions, leveraging prior knowledge, and adapting to dynamic instruction quality via implicit policy awareness. The toolkit comprehensively assesses distinct capabilities in structured data comprehension, function execution, instruction following, and critical reasoning. DrafterBench offers detailed analysis of task accuracy and error statistics, aiming to provide deeper insight into agent capabilities and identify improvement targets for integrating LLMs in engineering applications. Our benchmark is available at \href{https://github.com/Eason-Li-AIS/DrafterBench}{Github-DrafterBench}, with the test set hosted at \href{https://huggingface.co/datasets/Eason666/DrafterBench}{Huggingface}.
\end{abstract}

\section{Introduction}

Recently, Large Language Models(LLMs) have been demonstrated with remarkable capabilities in planning \citep{RN8}, problem-solving \citep{RN10}, tool calling \citep{RN12}, programming \citep{RN13}, etc. There is a growing trend for the integration of general LLMs in real scenarios, where one of the most promising fields is automating tasks by LLM agents in the industry \citep{RN7, RN58}. Such automation solutions are urgently needed in \textbf{Civil Engineering}, as there are bunches of monotonous, low-tech, and high-labor-intensity tasks from the construction stage \citep{RN22} to the design stage \citep{RN23}. They benefit users by helping them focus on more complex and skill-intensive work and create more value in the same work time. Systematic evaluation and comprehensive analysis from the perspective of industrial applications are critical to gain a deeper understanding of the capabilities of models and to identify targeted improvements. However, few benchmarks put them in the shoes of real-world industrial tasks, especially those of Civil Engineering.

\begin{figure}
    \centering
    \includegraphics[scale=0.60]{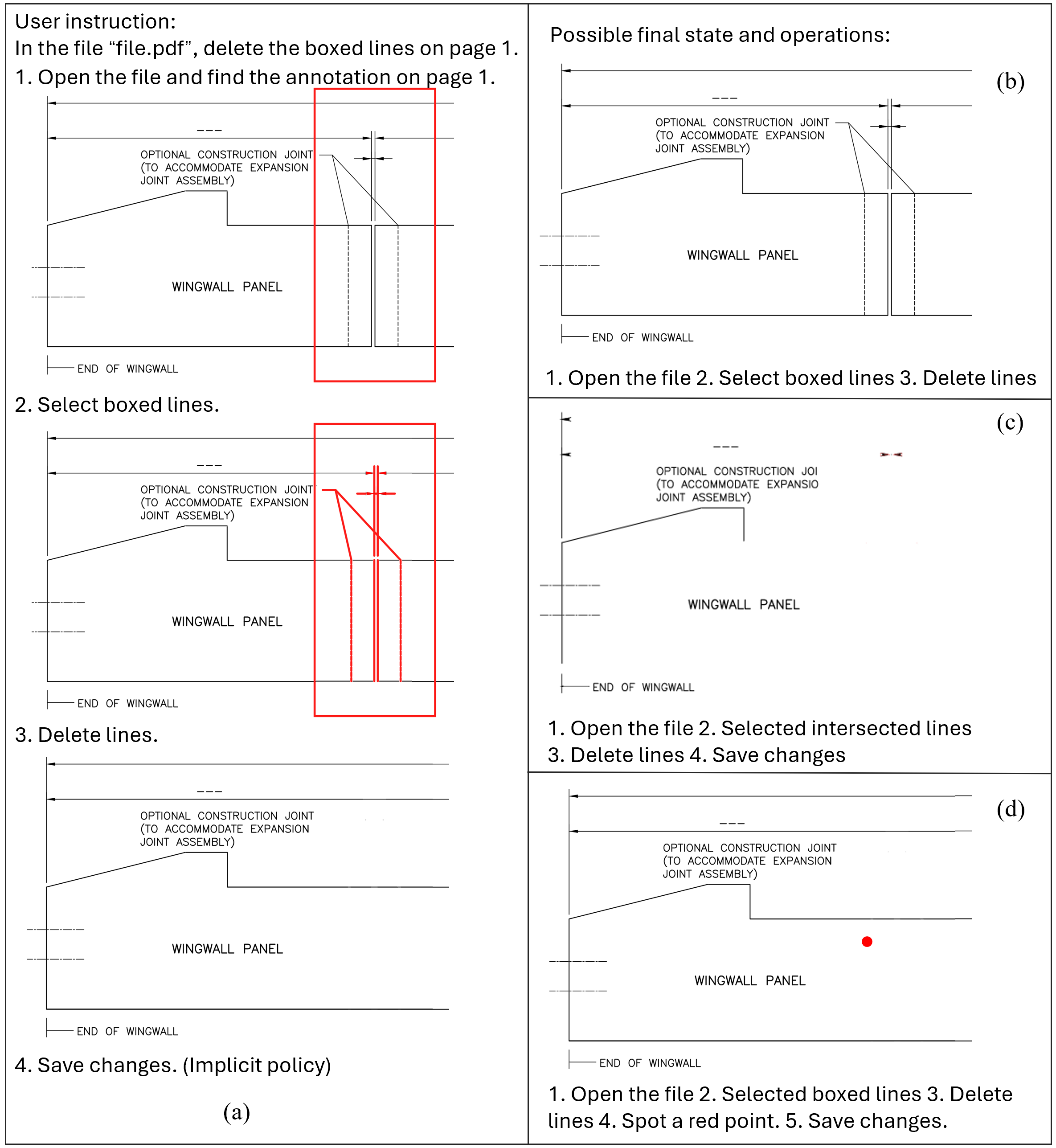}
    \caption{The workflow and possible results of a simple drawing revision task in civil engineering. (a) The whole process involves a series of operations, starting from opening the file, locating the annotation, selecting the expected lines, executing deletion, and ending with saving the changes. (b)(c)(d) illustrate the final outcomes of different WRONG operations.}
    \label{fig1}
    \vspace{-0.8cm}
\end{figure}

The innateness of industrial tasks brings unique challenges for both AI agents and benchmarks compared to those developed for general tasks. To illustrate them, we take the drawing revision as an example, which is one of the most labor-intensive and low-tech tasks that needs automation in Civil Engineering (according to our interviews with more than ten construction companies in North America). First, industrial tasks require a skilled worker to provide a complete solution by integrating available tools, prior knowledge, and implicit policies rather than simply calling functions following instructions. Figure \ref{fig1}a illustrates the complete process of performing a simple revision task. It is worth noting that after deleting the lines as instructed, the changes should be saved as a new file named according to the company or the user's policies. Usually, the prior knowledge and policies are common sense for a skilled human practitioner and are not mentioned in the instructions. But it is a challenge for LLMs who lack them to know them and act on them like a skilled worker. Second, high robustness is crucial in industrial tasks, which means that an exact drawing is expected even if an instruction is expressed in varying language styles and expressions from different individuals. Third, it is essential to ensure the accuracy of every detail in the workflow. A task fails even if a simple operation is omitted (forget to save changes, Figure \ref{fig1}b), a parameter value is in error (delete all lines rather than only boxed lines, Figure \ref{fig1}c), or an unexpected parameter is specified (an extra point appears, Figure \ref{fig1}d). Fourth, it is difficult to assess the quality of the LLMs' performance directly from the revised drawings. This is because not all steps in a solution make any visible changes, resulting in 'what can be seen is not what has been done'. It is possible that an unclean solution with invisible extra operations outputs a drawing that is the same as the ground truth.

Existing benchmarks actively explore the intelligence limit of LLMs in various scenarios \citep{RN34, RN33, RN35, RN36}, but they fall short of automating industrial tasks. They assess the capabilities of LLMs with the tasks of getting answers by following instructions straightforwardly and paying less attention to implicit policies. Besides, the interactive style of multi-turn and multi-round is a trend to evaluate the serviceability of agent assistants in the real world \citep{RN37, RN39}. However, agents in these styles are sometimes found to be unstable and overly creative, which conflicts with the high-robustness requirement in industrial tasks \citep{RN12, RN57, RN41}. Meanwhile, they usually require an observable environment or human feedback to promote the mandate process. However, for autonomous tasks in industry, building an observable environment may not be easy, and it is desirable to minimize human involvement to maximize their productivity. In addition, many benchmarks shed light on the completion of the final result, obscuring the comprehensive analysis of where a failure stems from \citep{RN31}. The latter is hugely helpful for having a deeper and clearer insight into the capabilities of models. This is especially true given that in industry tasks, ground truth results can be obtained both through ground truth paths in various forms and contaminated paths in accidents.

In this work, we introduce DrafterBench, an open-source automatic toolkit, to evaluate AI agents in automating a representation task of civil engineering, drawing revisions. By analyzing real drawing revision documents, 12 types of tasks are summarized across four types of operations (adding, content modification, mapping, and format updating) and three objects (text, table, and vector). Forty-six tools for revising drawings in PDF files are customized and provided with the necessary prior knowledge and implicit policy in the system information. As a result, 1920 tasks in dynamic instruction quality were prepared and verified by humans to simulate the real scenario. DrafterBench assesses four essential capabilities of models: structured data understanding, function execution, instruction following, and critical reasoning. It offers a systematic analysis of task accuracy and error statistics on six subtasks. To accurately assess the models, dual functions are designed to record ground operation paths, which may be expressed in various coding styles by different models in different responses. The operation paths, instead of the output drawings, were compared with the ground truth paths to grade the performance of the models and analyze detailed errors. Our contributions are as follows:

\begin{itemize}
  \item We introduce DrafterBench, which provides a comprehensive analysis of the strengths and limitations of LLM agents to automate monotonous and low-tech tasks for industry scenarios, especially civil engineering.
  \item A fully automated evaluation toolkit is released, providing a stable and accurate evaluation on models, resisting stochastic variation in models' response style and manner, avoiding the instability that may be encountered in interactive tasks.
  \item We conducted experiments for different mainstream LLMs. The results demonstrate that DrafterBench can give a clear view of their strengths and deficiencies. We hope that the proposed benchmark can provide insight for the future development of LLM agents, especially for integrating LLMs in engineering applications.
\end{itemize}

\section{Related Works}

\paragraph{LLM-based Agent}

Significant advances in LLM agents have been made with the development of the reasoning enhancement framework for LLM \citep{RN5, RN6, RN15, RN1}. By accomplishing relatively simple subtasks one by one, agents can solve complex tasks \citep{RN46, RN11}. ReAct is the most popular agent construction approach, which commands the agent to observe the environment to think deeply, complete a simple operation each turn, and gradually move toward the final answer \citep{RN15}. Based on the structure, agents can be categorized as single-agent or multi-agent. A single-agent accomplishes observing, planning, selecting, executing, and other actions by self-asking and answering \citep{RN53}, while a multi-agent system has independent modules responsible for each of the above four or more aspects and accomplishes the task through inter-module communication \citep{RN56}. Despite the ReAct-style agent's power, in industrial application scenarios, there is likely no environment for interaction, and agents that can accomplish tasks in a single turn are preferred.

\paragraph{Function-calling Benchmarks}

There are a number of benchmarks that have been built to evaluate the performance of agents, especially function-calling agents \citep{RN57, RN30, RN59}, on basic capabilities such as understanding, reasoning, planning, problem-solving, and their behavior in real tasks \citep{RN58, RN72}. The benchmarks have included a variety of tasks such as math, games, science quizzes, code, etc \citep{RN37, RN35}. However, few benchmarks have been built on the real engineering tasks that need to be automated. Meanwhile, recent benchmarks have focused chiefly on multi-step, multi-turn problem solving \citep{RN62, RN37} and not so much on one-turn, long-sequence high-level function callings, which are much more efficient in some tasks in the industry.

\section{The DrafterBench}

\begin{figure}
    \centering
    \includegraphics[scale=0.45]{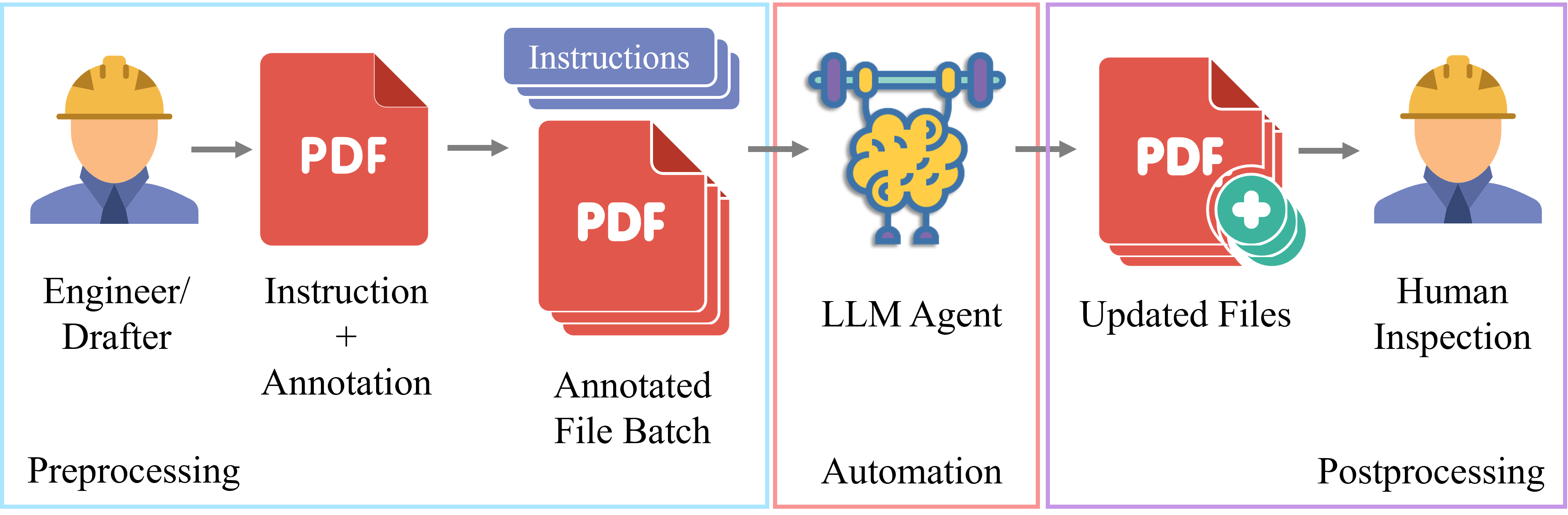}
    \caption{Expected workflow of an automation agent for drawing revision: The engineers or drafters leave their modification instructions and mark the related locations in the target files. Then, preprocessing will extract instructions and prepare files containing only the position marker for the agent. The LLM agent receives the instructions and files and automatically takes action to execute the modification. Finally, the workflow ends with a human inspection of the updated files.}
    \label{fig2}
    \vspace{-0.5cm}
\end{figure}

DrafterBench is designed to evaluate LLM models when serving as an automation agent to save engineers’ or drafters’ effort on low-tech monotonous work in industrial scenarios like drawing revision. Thus, tasks in DrafterBench are limited to the part of the work that can be completed by calling functions according to instructions and implicit policies. From a practical application perspective, this assumed automation agent has a workflow shown in Figure \ref{fig2}. This benchmark skips the preprocessing process and simulates the situation in which the agent receives the extracted instruction and prepared files and starts to take action. The action is calling the tools/functions provided by coding (which is more flexible and stable than other function calling methods \citep{RN73}) to implement the revision instruction in one turn without human participation. 

\subsection{Task Collection}

Over 100+ real-world drawing revision files (provided by design firms and construction companies) were collected and comprehensively analyzed. The target tasks were filtered out and categorized into three elements: text, table and vector entity, and four operations: adding, content modification, mapping, and format updating, in total 12 types. Detailed descriptions of them can be found in Appendix B. The difficulty of each task varies greatly and is generally controlled by six parameters in four dimensions:

\paragraph{Difficulty in understanding structured data}  Since these documents come from different companies and are handled by different people, the language style or expression of the instructions varies greatly, leading to a dynamic difficulty in understanding embedded data. They can be either extremely clear and concise sentences, which are named structured language in this work, or verbose paragraphs providing more and sometimes abundant information, namely unstructured language.

\begin{figure}
    \centering
    \includegraphics[width=\textwidth]{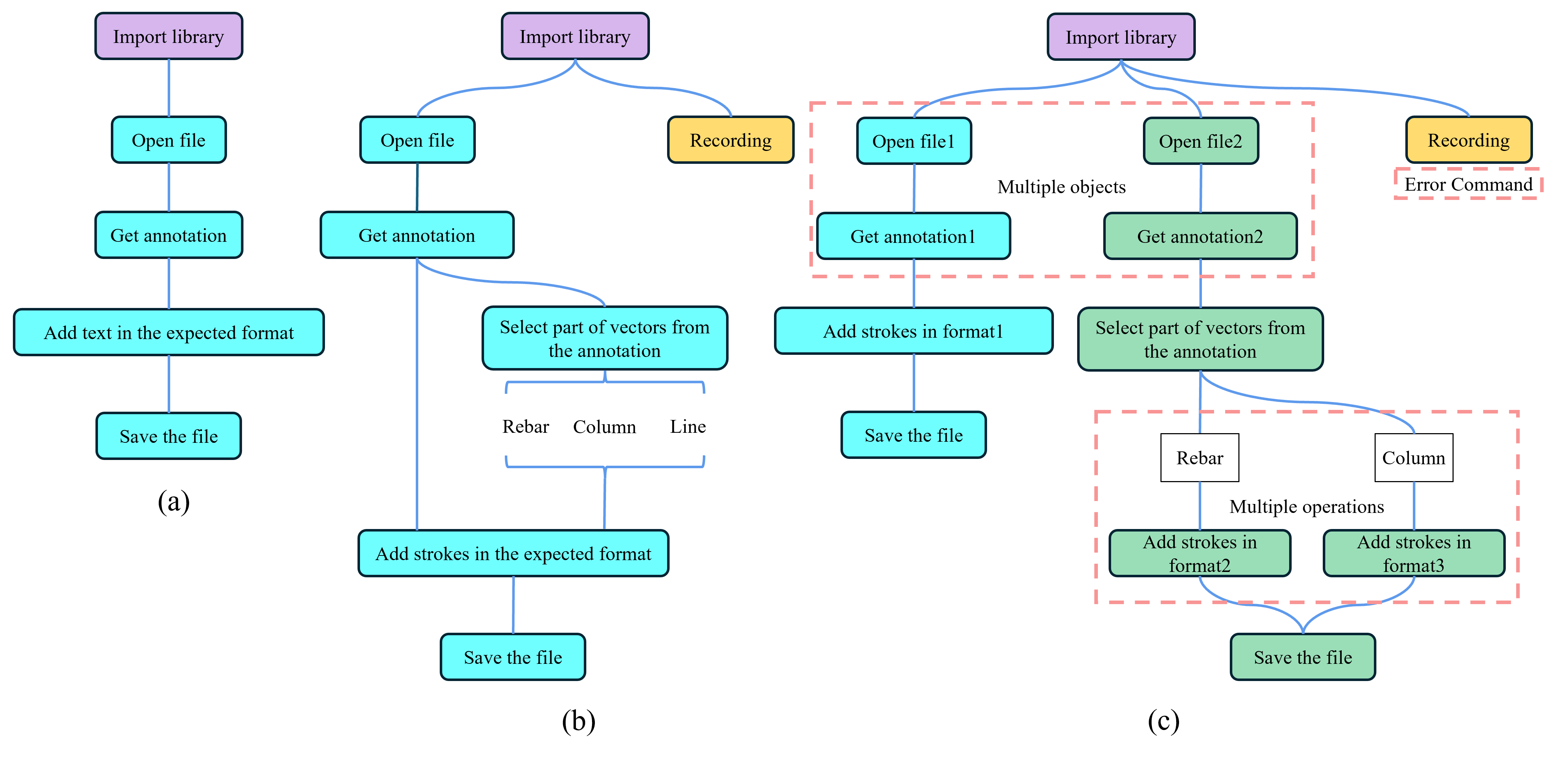}
    \caption{An illustration of function calling pipelines for different situations: (a) A standard pipeline for adding texts. (b) A standard pipeline for adding vectors. (c) Execution details for an instruction with multiple objects and long operation chains in one turn.}
    \label{fig3}
    \vspace{-0.5cm}
\end{figure}

\paragraph{Difficulty in function execution}  There are different function calling pipelines for each type of task. But their complexity varies greatly. Figure \ref{fig3}ab illustrates the pipelines for adding text and adding vectors, respectively, as examples. It can contain only straightforward steps in relatively easier tasks, like adding text, or a graph structure with richer nodes and edges for tasks like adding vectors.

\paragraph{Difficulty of instruction following}  The number of objects and revisions involved makes the length and complexity of an instruction fluctuate greatly. The impact of multiple objects and multiple revision operations on the implementation of an instruction is shown in Figure \ref{fig3}c. 

\paragraph{Difficulty of critical reasoning}  Due to the stochastic nature of human manual work, the quality of instructions is not always maintained at a constant standard. There are two types of errors that frequently appear. Sometimes, detailed values are not clearly specified, like “Move the table to the left a little”, that is, vaguely defining the value. It also happens that some necessary information is not specified, such as “add the missing dimension” without specifying the content or value to be added, which is an incomplete(error) instruction. The fully automated workflow expects the agent to respond to the above errors in two ways. The agent should think carefully and self-correct the ambiguous details with reasonable values. It is better to record incomplete instructions in the logging files using specified tools to alert the user rather than implement them anyway and make things worse.  

The six parameters are summarized in Table \ref{Task Specifications}. When constructing the benchmark, all filtered instructions were first grouped according to the six parameters. Tasks were sampled directly for groups with adequate tasks as candidate tasks or supplied with synthesized instructions for groups with fewer tasks. Then, the candidate tasks were verified by humans to ensure the solvability of the instructions and the uniformity of the value distribution. As a result, 1920 tasks were obtained with five tasks for each combination of difficulty parameters, as shown in Table \ref{Task Specifications}. 

\begin{table}[h]
	\caption{Task Specifications}
	\label{Task Specifications}
	\centering
	\begin{tabular}{llll}
		\toprule
		Items                      & \multicolumn{2}{l}{Parameter   details}    & Number \\
		\midrule
		Task category                & \multicolumn{2}{l}{12 types of tasks}                    & 12     \\
		Object per instruction       & \multicolumn{2}{l}{Single object/Multiple objects}         & 2      \\
		Operation per instruction    & \multicolumn{2}{l}{Single operation/Multiple operations} & 2      \\
		Information completeness & \multicolumn{2}{l}{Complete/Incomplete} & 2      \\
		Value specifying style & \multicolumn{2}{l}{Precisely/Vaguely}                      & 2      \\
		Language style           & \multicolumn{2}{l}{Structured/Unstructured}                & 2      \\
		Tasks per situation      &                     &                        & 5      \\
		Total                      &                     &                        & 1920   \\
		\bottomrule
	\end{tabular}
\end{table}

\subsection{Tool Preparation}

\paragraph{Drawing revision tools/functions}
The tools/functions provided for the agent were specifically tailored for PDF drawing revision in this work, a total of 46 tools. The well-known PDF editing libraries, including PyMuPDF \citep{RN42} and Reportlab \citep{RN43}, and computer vision libraries, such as cv2 \citep{RN44} and pyresseract \citep{RN45}, are introduced when designing them.

\paragraph{Dual tools/functions}
As it is difficult to conduct an accurate evaluation of the performance of models directly from the output drawings, dual tools/functions are introduced. They have function names, calling methods, arguments, and output types the same as the corresponding original functions. When the agent executes the generated code, the dual functions are executed instead of the original functions. They do not modify the drawing but record the argument details and the ground operations path, following the rules, such as repeated operations being recorded as one. They allow and record common coding errors that are not allowed in the original functions, such as parameter type errors. These records will be compared with the ground truth for analysis at the operational level. It can clearly estimate the operation quality, eliminate the noise introduced by coding styles, and easily distinguish unclean paths even though they have output drawings that are the same as the ground truth paths. In summary, the dual tools facilitate an accurate and comprehensive assessment of the mode while ensuring the uniqueness of the ground truth and considering the flexibility of the coding.

\subsection{Default Prompt}

After several experiments and adjustments, we have obtained a prompt framework that can effectively solve the tasks of drawing revision by coding with function calling, following the instructions and implicit policies in this benchmark. The default prompts for each task can be found in Appendix D. We also release the editing privileges of the prompts to encourage users to develop their own prompts to achieve higher performance.

\section{Evaluation Metric} \label{eval}

The evaluation metric was developed to evaluate both the task accuracy and the error statistics of the agent's performance. The operation paths recorded by dual functions are compared with the ground truth paths. For each task, its score is graded by three parts: code executability, target completeness (completeness of the instruction).

The evaluation has two levels: \textit{Level 1}, check whether a generated code can be run with dual functions. The dual functions are compatible with some common coding errors. Thus, there must be significant errors in an unrunnable code string and it will receive a score of 0. \textit{Level 2}, when a code string can be run with dual functions. Each function execution is monitored for executability and will be recognized as executable if and only if (a) all required arguments are specified and (b) all specified arguments are in the correct data format. The executability score of the response is graded as 0 if any function is non-executable; otherwise, it is 30. The target completeness is assessed in six subtasks described below.

\begin{wrapfigure}{r}{0.5\textwidth}
	\centering
	\includegraphics[width=0.5\textwidth]{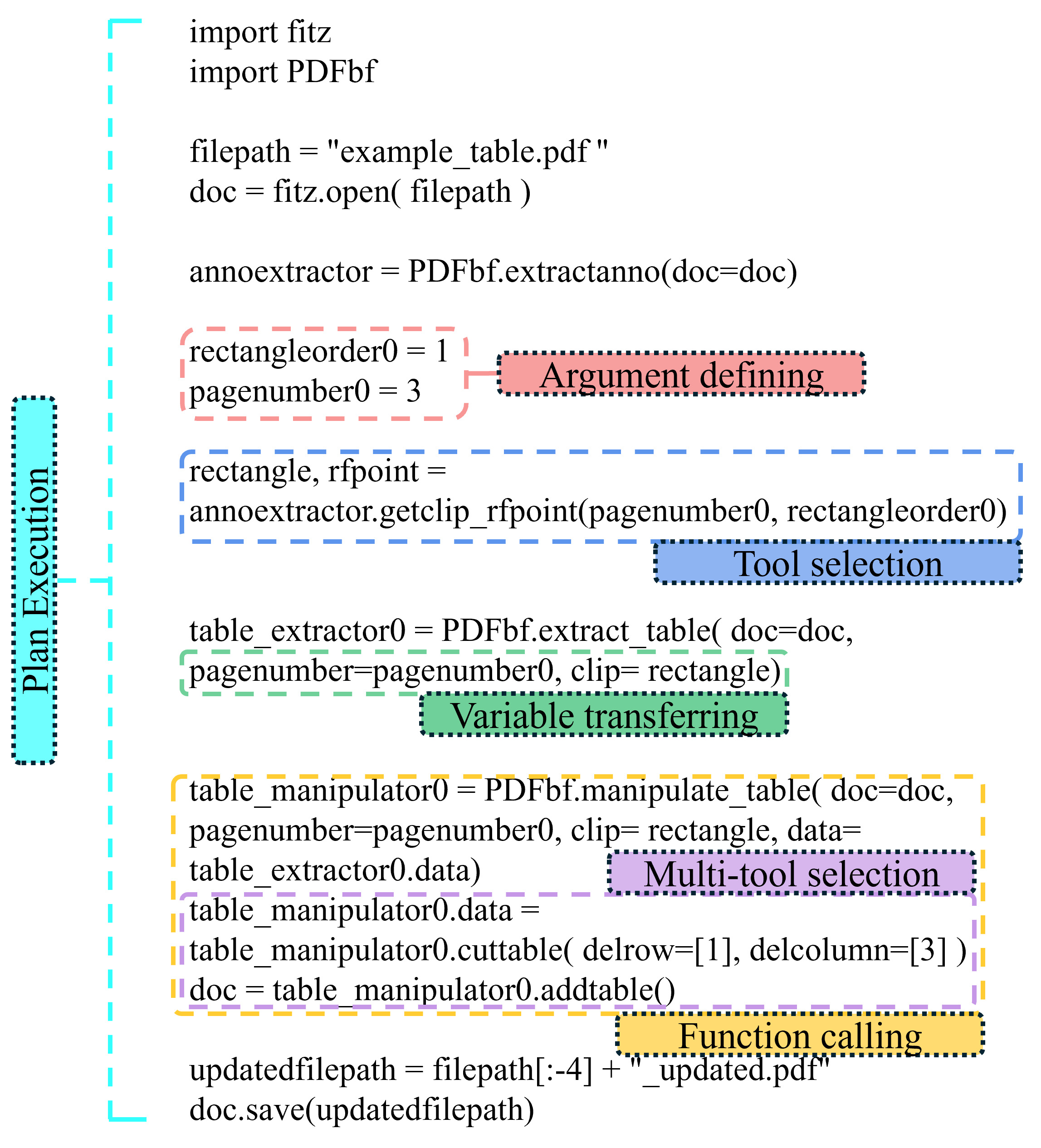}
	\caption{Subtasks when calling a function by coding to revise a drawing.}
	\label{fig4}
\end{wrapfigure}
\textbf{Argument defining:} define arguments according to the details in the instruction.\\
\textbf{Variables transferring:} transfer intermediate variables between functions.\\
\textbf{Function calling:} call functions or tools.\\
\textbf{(Single) Tool selection:} select a suitable tool from the tool library.\\
\textbf{Multi-tool selection:} select a series of tools in sequence from the tool library.\\
\textbf{Plan execution:} The critical details of a sequence of functions corresponding to an object's revision operations.\\
Figure \ref{fig4} illustrates the six subtasks. The scores for the first five subtasks were calculated using Equation \ref{XX}. 
\begin{equation}
		\frac{TP-FP}{TP+FN}\times\frac{70}{6} \label{XX} \\
\end{equation}
\begin{equation}
	\sum _{\left (p^{p}, p^{g}\right )} IoU\left ( p^{p}, p^{g} \right ) \times \frac{70}{6} \label{YY} \\
\end{equation}
\begin{equation}
	mean\left ( s_{i} \right ) - \frac{\left ( 100-min\left ( s_{i} \right )  \right ) }{n}  \label{ZZ} \\
\end{equation}

\textit{TP} refers to the number of subtasks in a response that are the same as the ground truth, and vice versa for \textit{FN}. \textit{FP} is the number of subtasks that are not in the ground truth, but are performed in the response. The drawing revision task is a result-sensitive task. If an agent performs unexpected operations, they are likely to introduce new errors and make things worse. Therefore, \textit{FP} is introduced as a penalty when scoring the response. 

The intersection over Union (IoU) was employed to score the plan execution. \textit{p} is a sequence of revision operations performed on an object. $p^g$ refers to the ground truth, and $p^p$ refers to the response of the agent. Equation \ref{YY} was employed to score plan execution. 
The total score for all subtasks is 70. The sum of executability and target completeness score is the total score of the agent's response to an instruction.

The comprehensive score of the agent is calculated by Equation \ref{ZZ}, where $s_i$ is the average score for the \textit{i}-th type of task. The second part is introduced to better consider the weak points of the agent. If we simply take the average score of the 12 tasks as the comprehensive score, it is impossible to distinguish between the following two situations: 1) Agents scored high in some sets and low in others. 2) Agents scored evenly on each type of task. These two situations may have similar comprehensive scores, but the latter is preferred in the industry scenario.

\section{Experimental Results}

\begin{table}[]
	\caption{Average task scores for different task sets and results of the comprehensive evaluations. Bold and underlined numbers are the first and second-highest performances in each category.}
	\label{Tab2}
        \large
	\resizebox{\textwidth}{!}{
	\begin{tabular}{>{\centering\arraybackslash}m{0.15\textwidth}
			>{\centering\arraybackslash}p{0.15\textwidth}cccccc}
		\toprule
		\multicolumn{2}{l}{}                              & OpenAI o1 & ChatGPT-4o-2024-08-06 & Claude3.5-sonnet & Deepseek-v3-685B & Qwen2.5-72B-Instruct & Llama3-70B-Instruct \\
		\midrule
		\multirow{2}{0.15\textwidth}{Language style}                   & Structured language   & \textbf{81.58} & 75.14      & 74.34       & {\ul 76.26}  & 75.31          & 70.71           \\
		\cmidrule{2-2}
		& Unstructured language  & \textbf{82.26} & 73.84      & {\ul 78.20}      & 75.84       & 74.23          & 70.62           \\
		\cmidrule{1-2}
		\multirow{2}{0.15\textwidth}{Details ambiguity}       & Precise details     & \textbf{89.82} & 79.46      & {\ul 81.15}      & 79.25       & 76.18          & 73.50           \\
		\cmidrule{2-2}
		& Vague details & \textbf{74.02} & 69.52      & 71.39            & {\ul 72.86} & 73.36          & 67.84           \\
		\cmidrule{1-2}
		\multirow{2}{0.15\textwidth}{Instruction   completeness}                & Complete instruction      & 81.43          & 79.63      & 82.00            & {\ul 82.74} & \textbf{84.49} & 80.96           \\
		\cmidrule{2-2}
		& Incomplete instruction          & \textbf{82.41} & 69.36      & {\ul 70.54}      & 69.37       & 65.06          & 60.38           \\
		\cmidrule{1-2}
		\multirow{2}{0.15\textwidth}{Objects number}                   & Single object          & \textbf{83.23} & 73.87      & {\ul 75.36}      & 75.28       & 75.07          & 69.57           \\
		\cmidrule{2-2}
		& Multiple objects       & \textbf{80.60} & 75.11      & {\ul 77.19}      & 76.83       & 74.47          & 71.77           \\
		\cmidrule{1-2}
		\multirow{2}{0.15\textwidth}{Maximum operation   chain length} & Single operation       & \textbf{82.06} & 75.46      & 76.15            & {\ul 77.21} & 76.26          & 72.52           \\
		\cmidrule{2-2}
		& Multiple operations    & \textbf{81.63} & 72.56      & {\ul 76.52}      & 73.74       & 71.79          & 66.97           \\
		\cmidrule{1-2}
		\multicolumn{2}{c}{Average tasks score}                         & \textbf{81.92} & 74.49      & {\ul 76.27}      & 76.05       & 74.77          & 70.67           \\
		\cmidrule{1-2}
		\multicolumn{2}{c}{Comprehensive score}                 & \textbf{79.90} & 71.76      & {\ul 73.79}      & 73.09       & 72.05          & 67.55 \\
		\bottomrule  
	\end{tabular}
}
\end{table}

We test various state-of-the-art commercial and open-source language models for agents through their APIs: OpenAI GPT API (o1, gpt-4o-2024-08-06) \citep{RN64}, Anthropic Claude API (claude-3-5-sonnet-20241022) \citep{RN66}, Deepseek API (deepseek-chat) \citep{RN67}, Deepinfra API (Qwen/Qwen2.5-72B-Instruct, Meta-Llama-3-70B-Instruct) \citep{RN68, RN65}.

\subsection{Task Accuracy and Robustness}

Table \ref{Tab2} illustrates the main results. First, OpenAI o1 leads the performance in almost all task sets with obvious dominance compared to other models. There are two models in the second tier, Claude3.5-sonnet and Deepseek-v3-685B, which are neck-and-neck for the second-highest performance. Second, even the widely recognized powerful OpenAI o1 fails to earn around 20 points for simple, monotonous, and low-tech industrial tasks, highlighting the necessity of evaluation centered on industrial applications. Third, the performance of all models fluctuates in tasks with different difficulty, especially for tasks with some sensitive parameters.

\begin{figure}
	\centering
	\includegraphics[width=\textwidth]{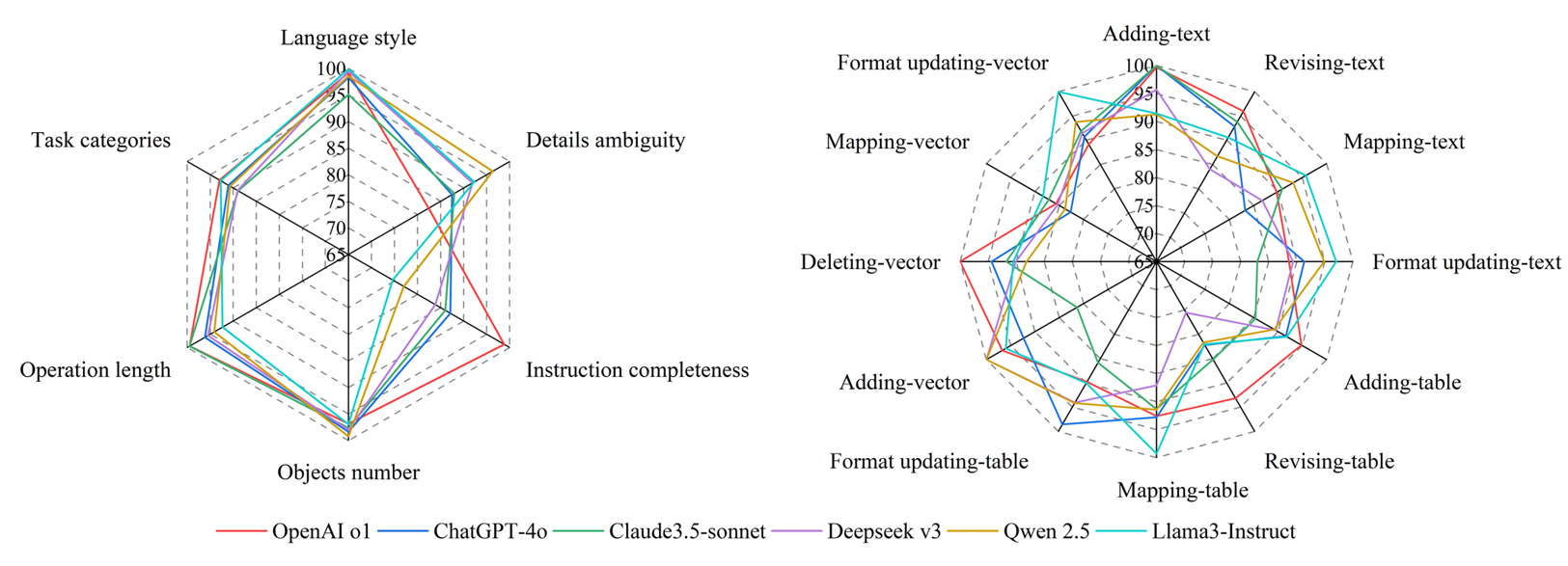}
	\caption{An illustration of performance degradation in terms of six parameters (left) and detailed performance of each task type (right). The values in the graph are the percentage of the model's performance in the weaker task set versus its performance in the stronger task set. }
	\label{fig5}
        \vspace{-0.5cm}
\end{figure}

To dissect the capabilities of LLMs to automate drawing revision tasks and to gain insight into their robustness, the agent's performance in tasks of different complexity is shown in Figure \ref{fig5}. From the robustness of \textbf{structured data understanding} in different language styles, almost all models show great stability with an average degradation of 1\%, except for the Claude3.5-sonnet (5\%). For the robustness of \textbf{function execution}, the performance degradation of each type of task versus the best type of task for each model is around 9\%. No significant gap was observed between the models. The relatively most stable model, OpenAI o1, decreases by 7\%, and the least stable model, Claude3.5-sonnet, decreases by 11\%. There is an interesting observation in Figure \ref{fig5} that the best performance occurs in tasks having the least number of arguments (deleting vectors – o1, adding vectors – Deepseek-v3-685B, Qwen2.5-72B-Instruct) or having the most simple pipeline (adding text – ChatGPT-4o-2024-08-06, Claude3.5-sonnet, updating vector format - Llama-3-70B-Instruct). From the robustness of \textbf{instruction following}, the good news is that all models exhibit good stability with a performance degradation of less than 4\% when dealing with tasks involving multiple objects compared to only involving a single object. Meanwhile, when dealing with tasks having multiple operations, OpenAI o1 and Claude3.5-sonnet exhibit excellent stability with a degradation of less than 1\%, while that of others is 6\% on average. From the robustness of \textbf{critical reasoning}, all models somehow struggle to complete the tasks. When details are vaguely specified, the degradation of all models is great, 5\% for Qwen2.5-72B-Instruct and 12\% for others. When there are errors in the instructions, performance degradation reaches 18\% for all models except OpenAI o1, which shows great resilience, in contrast.

\subsection{Error Analysis}

\begin{wrapfigure}{r}{0.5\textwidth}
	\centering
	\includegraphics[width=0.5\textwidth]{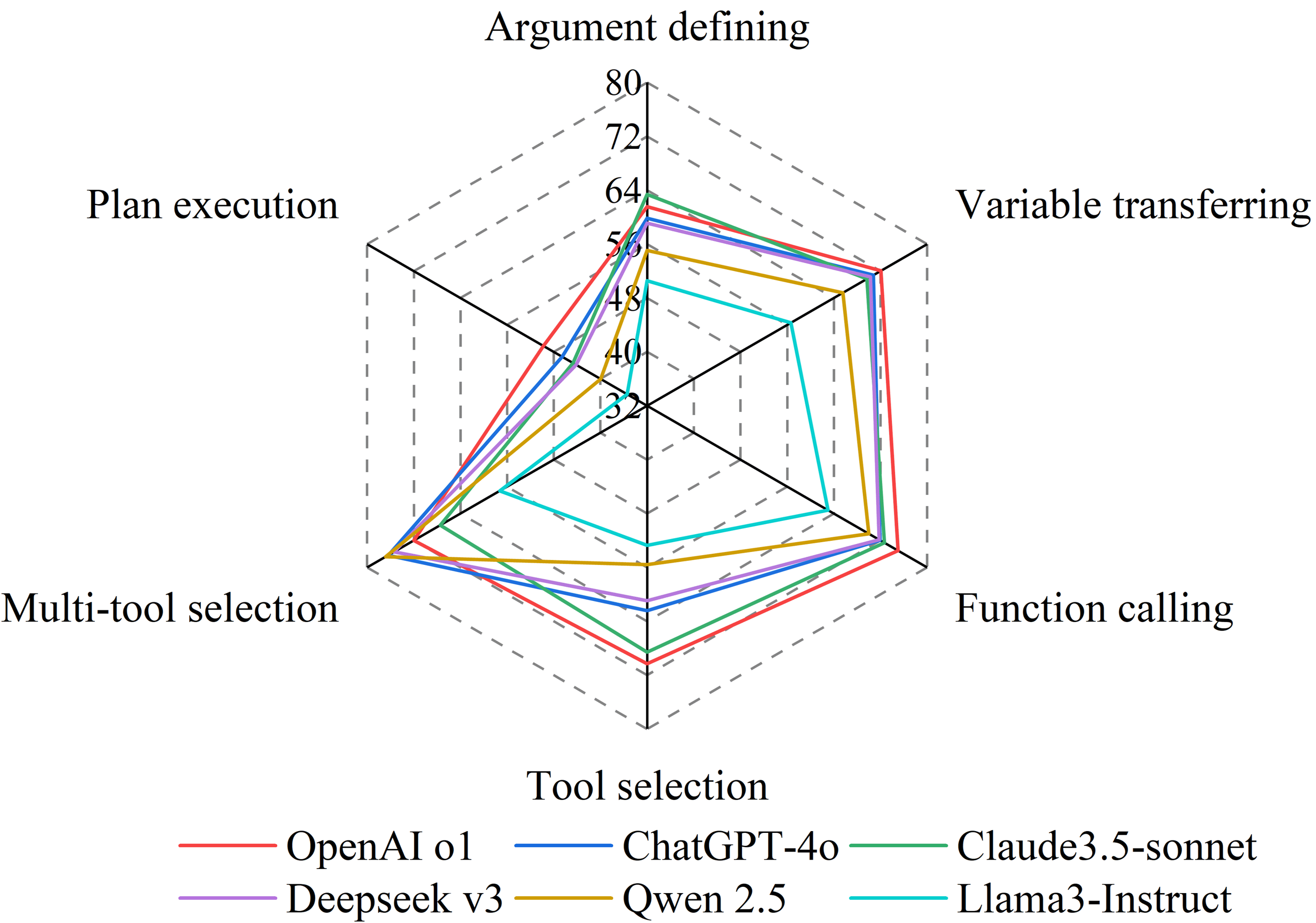}
	\caption{Subtasks accuracy(\%) of tested models.}
	\label{fig6}
\end{wrapfigure}
To have a clear understanding of where the errors stem from and to have a deeper understanding of the limitations of the models, we calculated the accuracy of the six subtasks mentioned in Section \ref{eval}, as shown in Figure \ref{fig6}. Similar to the comprehensive score, OpenAI o1 leads the way in accuracy in almost all subtasks, followed by Claude3.5-sonnet and Deepseek-v3-685B.
To benefit the reader's reading experience, only a comprehensive analysis is provided here; a more detailed discussion, such as how the performance of six subtasks varies at task sets in different difficulties, can be found in Appendix C. 

In Figure \ref{fig6}, there is an obvious gap between the accuracy of plan execution and the other five subtasks. It is interesting that the difference is constantly around 20\% for all models, though the value of their subtask accuracy varies greatly. OpenAI o1, one of the most advanced models in the world, shows the same gap, even though it exceeds other models in almost all other subtasks. The plan execution was scored by checking whether all critical details of a sequence of tools to modify an object were consistent with the ground truth. Therefore, it can be deduced that the models are trying their best to complete each subtask with high accuracy, but it is difficult for them to pay enough attention to every detail. The models are able to understand a significant portion of the task objectives, but still lack the ability to avoid all errors throughout the modification process of an object.

\subsection{Limitations and Future Directions}

To fulfill the requirements of automating monotonous, low-tech, and high-labor-intensity tasks in the industry, like drawing revision tasks, there are still several challenges that need to be considered in the future development of LLMs.

\paragraph{Interactive style}
In our experiments, when LLMs encounter an incomplete instruction, they prefer to ask the user for more information or to fill the missing information with a placeholder to give the output rather than record it according to the polices in the system information. This contributes to the low accuracy in the incomplete instruction task set. It also corroborates the fact that current LLMs are paranoid about instantaneous interaction with humans. Instantaneous interaction is indeed an important scenario for the application of LLMs. But in real engineering applications, especially when automating industrial tasks, there are also scenarios that prefer no human involvement, such as the drawing revision task considered in this benchmark. Although LLMs are evolving toward super-intelligence and seamless interaction with humans, it is also important to maintain compatibility with other modes of interaction, such as delayed, indirect interaction through preprocessing and postprocessing files between LLMs and humans.

\paragraph{Understanding of details}  
LLMs are in trouble when dealing with instructions that vaguely define key arguments. Common errors include filling in placeholders or directly filling in the description words in the instructions for vaguely defined arguments. Therefore, it is necessary to strengthen LLMs to speculate, reason, and understand the user's exact intention behind the instruction details rather than literally executing them and giving a response for reference only. 

\paragraph{Implementation of new policies}  
LLMs perform quite stubbornly when they meet instructions that are in some kind of conflict with their intrinsic policy. For example, during the testing of this benchmark, the system information in the prompt repeatedly emphasizes the user's expected response when encountering detail-ambiguous instructions and error instructions. They can obey the police in some tasks, but also ignore the new policies in a significant number of subtasks and give the unexpected responses mentioned above, in accordance with their habits or intrinsic policy. Having and remaining loyal to intrinsic policies is a necessary competency for LLMs, especially when it comes to cybersecurity or illegal risk. However, in some other less serious and stereotypical aspects, there is a need to improve LLMs' tolerance and implementation of new policies, especially considering the diverse demands and polices of industrial application scenarios. More benchmarks are also needed to illustrate more requirements for integrating LLMs in industrial scenarios.

\section{Conclusion}

In this work, we introduce DrafterBench, an open-source toolkit designed to offer a systematic and comprehensive evaluation of the capabilities of LLMs to automate monotonous, low-tech, and high-labor-intensity tasks from industry. We emphasize the challenges and rigorous requirements for AI agents from an industrial perspective. DrafterBench is under the context of a representation civil engineering task, drawing revision, a total of 1920 drawing revision tasks are collected from real documents. The complexity of tasks is controlled by six parameters to investigate four essential capabilities of LLMs, namely structured data understanding, function execution, instruction following, and critical reasoning. The automatic evaluation toolkit assesses the performance of LLMs with task accuracy and error statistics. We conducted experiments on mainstream LLMs to reveal their strengths and deficiencies in automating industry tasks. From the result, we posit that DrafterBench is a useful metric that can provide valuable evaluation and useful insight for the future development of LLMs from the perspective of industry, especially Civil Engineering.

\clearpage

\newpage
\appendix

\section{Limitation and Computing Resources}

\subsection{Limitation}
The tasks involved in this work are all in English. Multilingual input will be included in future work to consider the ability of LLMs to automate industrial tasks in non-English environments. At the same time, this work covers only drawing revision tasks in the field of civil engineering, and more types of tasks are yet to be performed in future work.

\subsection{Computing Resources}
This benchmark calls LLMs via APIs and has no special requirements for hardware devices. In this experiment, the benchmark runs on an i9-13900F CPU for about 30 min to get all the results.

\section{Description of Drawing Revision Tasks}
The detailed description of the 12 drawing revision tasks is shown in Table \ref{12 tasks}.

\begin{table}[h]
	\caption{12 types of tasks for the drafter agent.}
	\label{12 tasks}
	\centering
	{
		\begin{tabularx}{\textwidth}{m{3cm}m{3cm}X}
			\toprule
			Object element                   & Operation       & Description                                                                                                                  \\
			\midrule
			\multirow{4}{*}{Text}            & Adding          & Add new texts to the target area.                                                                                            \\
			& Content modification        & Delete or replace words in the target text strings.                                                                        \\
			& Mapping         & Move, rotate or scale the target texts.                                                                                     \\
			& Format updating & Modify the format of target texts, including but not limited to text color, font type, font size, and alignment.      \\ 
			\midrule
			\multirow{4}{*}{Table}           & Adding          & Add new tables to the target area.                                                                                          \\
			& Content modification        & Delete, clear, or replace cells in the target tables.                                                                       \\
			& Mapping         & Move, rotate or scale the target tables.                                                                                    \\
			& Format updating & Modify the format of target tables, including but not limited to font type, font size, boundary width, and alignment. \\ 
			\cmidrule(r){1-3}
			\multirow{4}{*}{Vector entity} & Adding          & Add standard strokes to the target position.                                                                                 \\
			& Content modification        & Delete target strokes, including rebars, columns, and lines.                                                           \\
			& Mapping         & Move, rotate or scale target strokes.                                                                                   \\
			& Format updating & Modify the format of target strokes, including but not limited to line color,   line thickness, and line type.         \\  
			\bottomrule
		\end{tabularx}
	}
\end{table}

\section{Performance of Agents for Tasks in Different Difficulty}

\subsection{Structured Data Understanding}

\begin{table}[h!]
	\caption{Accuracy of subtasks in different language styles in percentage. STR means structured language, and U-STR means unstructured language.}
	\label{tab4}
	\resizebox{\textwidth}{!}{
	\begin{tabular}{lllllllllllll}
		\toprule
		& \multicolumn{2}{c}{OpenAI o1} & \multicolumn{2}{c}{ChatGPT-4o-2024-08-06} & \multicolumn{2}{c}{Claude3.5-sonnet} & \multicolumn{2}{c}{Deepseek-v3-685B} & \multicolumn{2}{c}{Qwen2.5-72B-Instruct} & \multicolumn{2}{c}{Llama3-70B-Instruct} \\ 
		\cmidrule{2-13} 
		& STR           & U-STR         & STR            & U-STR         & STR               & U-STR            & STR            & U-STR          & STR           & U-STR        & STR              & U-STR            \\
		\midrule
		Argument defining       & 59.84         & 63.31         & 58.86          & 60.81         & 62.81             & 63.94            & 59.72          & 58.51          & 57.05         & 53.09        & 54.67            & 46.45            \\ 
		Variable transferring   & 71.03         & 73.13         & 70.95          & 70.62         & 72.67             & 66.70            & 72.92          & 67.64          & 68.44         & 62.68        & 63.36            & 49.97            \\
		Function calling        & 73.20         & 76.85         & 71.63          & 72.24         & 73.78             & 71.56            & 72.28          & 71.27          & 70.92         & 69.08        & 68.58            & 57.43            \\
		(Single) Tool selection & 68.31         & 72.28         & 63.28          & 61.57         & 67.30             & 69.87            & 62.58          & 59.31          & 57.44         & 53.72        & 52.72            & 52.72            \\
		Multi-tool selection & 70.15         & 73.86         & 75.38          & 77.40         & 76.22             & 58.68            & 75.89          & 74.20          & 77.91         & 75.72        & 77.07            & 37.10            \\
		Plan execution          & 49.94         & 49.59         & 48.09          & 45.09         & 47.11             & 42.45            & 45.29          & 43.07          & 41.64         & 38.27        & 39.57            & 31.75            \\
		\bottomrule       
	\end{tabular}
}
\end{table}

Table \ref{tab4} lists agents' performance for instructions in structured and unstructured language. The stable scores of all LLMs show their excellent adaptability to different language styles. The difference in both comprehensive scores and subtasks between the two language styles is around 5\%.

\subsection{Instruction Following}

\paragraph{Objects per instruction} The number of objects per instruction determines whether the pipeline will be implemented in parallel. As shown in Table \ref{tab5}, the good news is that LLMs perform stably for both a single object per instruction and multiple objects per instruction. An interesting observation is that models except OpenAI o1 perform better for multi-object instruction, which is actually more difficult. However, the increase in accuracy in all subtasks is not due to better performance but to a lower requirement for “Recording” incomplete instructions. For an incomplete instruction with a single object, the only ground truth is to call the “Recording” function; it will be 100\% false if it is not called. In contrast, for a multi-object instruction, “Recording” should only be called for objects that lack necessary information, and the standard pipeline should be executed normally for others. Therefore, LLMs can get higher scores even though none of the “Recording” is called in the multiple objects instructions.

\begin{table}[h]
	\caption{Accuracy of subtasks for instructions with a single object (SIN) or multiple objects (MULT).}
	\label{tab5}
	\resizebox{\textwidth}{!}{
		\begin{tabular}{lllllllllllll}
			\toprule
			& \multicolumn{2}{c}{OpenAI o1} & \multicolumn{2}{c}{ChatGPT-4o-2024-08-06} & \multicolumn{2}{c}{Claude3.5-sonnet} & \multicolumn{2}{c}{Deepseek-v3-685B} & \multicolumn{2}{c}{Qwen2.5-72B-Instruct} & \multicolumn{2}{c}{Llama3-70B-Instruct} \\
			\cmidrule{2-13} 
			& SIN           & MULT          & SIN            & MULT          & SIN               & MULT             & SIN            & MULT           & SIN           & MULT         & SIN              & MULT             \\
			\midrule
			Argument defining       & 62.65         & 60.93         & 53.77          & 63.40         & 60.72             & 64.93            & 50.70          & 64.05          & 46.25         & 60.24        & 34.89            & 59.76            \\
			Variable transfering   & 72.45         & 71.90         & 67.93          & 72.17         & 70.73             & 69.17            & 64.44          & 73.12          & 61.34         & 67.61        & 37.68            & 65.90            \\
			Function calling        & 76.64         & 74.13         & 68.22          & 73.98         & 73.95             & 71.97            & 66.04          & 74.94          & 65.13         & 72.68        & 47.08            & 71.78            \\
			(Single) Tool selection & 74.73         & 67.00         & 62.10          & 62.66         & 70.48             & 67.18            & 58.62          & 62.66          & 54.84         & 56.13        & 49.94            & 54.78            \\
			Multi-tool selection & 67.95         & 73.00         & 67.52          & 78.57         & 68.38             & 67.23            & 59.40          & 79.31          & 64.10         & 79.94        & 0.00             & 79.20            \\
			Plan execution          & 41.13         & 54.80         & 38.17          & 51.64         & 37.41             & 48.85            & 34.23          & 50.34          & 31.58         & 45.17        & 24.97            & 42.55            \\
			\bottomrule       
		\end{tabular}
	}
\end{table}

\paragraph{Operations per object} The number of operations per object changes the execution loop of sub-steps in the standard pipeline. The LLMs demonstrate some adaptability to varying operations. The difference between instructions that perform only one operation on an object and those that perform multiple operations is about 5\% for all subtasks. It is noted that some local drops also occur in multi-tool selection, as shown in Table \ref{tab6}.

\begin{table}[h]
	\caption{Accuracy of subtasks for instructions with a single operation per object(SIN) or multi-operations per object(MULT).}
	\label{tab6}
	\resizebox{\textwidth}{!}{
		\begin{tabular}{lllllllllllll}
			\toprule
			& \multicolumn{2}{c}{OpenAI o1} & \multicolumn{2}{c}{ChatGPT-4o-2024-08-06} & \multicolumn{2}{c}{Claude3.5-sonnet} & \multicolumn{2}{c}{Deepseek-v3-685B} & \multicolumn{2}{c}{Qwen2.5-72B-Instruct} & \multicolumn{2}{c}{Llama3-70B-Instruct} \\ 
			\cmidrule{2-13} 
			& SIN           & MULT          & SIN            & MULT          & SIN               & MULT             & SIN            & MULT           & SIN           & MULT         & SIN              & MULT             \\
			\midrule
			Argument defining       & 64.64         & 59.58         & 58.50          & 61.86         & 61.52             & 66.23            & 61.87          & 57.33          & 55.29         & 54.63        & 52.49            & 49.29            \\
			Variable transfering   & 71.20         & 72.74         & 72.51          & 68.47         & 67.96             & 72.00            & 67.77          & 72.15          & 68.76         & 61.28        & 57.02            & 56.40            \\
			Function calling        & 75.16         & 74.93         & 73.07          & 70.22         & 70.66             & 75.71            & 71.85          & 71.73          & 71.92         & 67.09        & 63.60            & 62.61            \\
			(Single) Tool selection & 70.14         & 70.41         & 64.14          & 59.91         & 69.81             & 66.79            & 59.54          & 61.90          & 58.51         & 51.30        & 48.39            & 55.67            \\
			Multi-tool selection & 72.71         & 71.57         & 76.79          & 75.76         & 59.62             & 73.36            & 77.73          & 73.90          & 77.88         & 75.11        & 68.34            & 36.54            \\
			Plan execution          & 53.45         & 47.26         & 44.51          & 49.42         & 41.93             & 48.82            & 46.37          & 42.55          & 38.56         & 41.92        & 39.63            & 32.75            \\
			\bottomrule       
		\end{tabular}
	}
\end{table}

\subsection{Critical Reasoning}

\begin{table}[b!]
	\caption{Performance of agents for instructions specifying values precisely (P)/vaguely (VA).}
	\label{tab7}
	\resizebox{\textwidth}{!}{
		\begin{tabular}{lllllllllllll}
			\toprule
			& \multicolumn{2}{c}{OpenAI o1} & \multicolumn{2}{c}{ChatGPT-4o-2024-08-06} & \multicolumn{2}{c}{Claude3.5-sonnet} & \multicolumn{2}{c}{Deepseek-v3-685B} & \multicolumn{2}{c}{Qwen2.5-72B-Instruct} & \multicolumn{2}{c}{Llama3-70B-Instruct} \\ 
			\cmidrule{2-13} 
			& P             & VA            & P              & VA            & P                 & VA               & P              & VA             & P             & VA           & P                & VA               \\
			\midrule
			Argument defining       & 84.53         & 38.87         & 69.88          & 49.91         & 72.12             & 54.72            & 65.94          & 52.38          & 57.43         & 52.72        & 51.16            & 49.96            \\
			Variable transfering   & 90.58         & 55.48         & 79.26          & 63.18         & 75.38             & 64.57            & 72.79          & 68.03          & 65.38         & 65.72        & 52.64            & 60.28            \\
			Function calling        & 93.15         & 58.17         & 79.46          & 64.94         & 76.72             & 68.91            & 75.46          & 68.35          & 69.99         & 70.01        & 58.88            & 66.84            \\
			(Single) Tool selection & 91.04         & 50.73         & 68.91          & 56.30         & 73.47             & 63.98            & 63.26          & 58.75          & 54.66         & 56.45        & 53.11            & 52.35            \\
			Multi-tool selection & 90.47         & 55.71         & 86.87          & 67.14         & 70.86             & 64.44            & 82.19          & 69.37          & 78.06         & 75.71        & 38.67            & 73.65            \\
			Plan execution          & 63.14         & 38.39         & 54.90          & 38.59         & 51.04             & 38.17            & 49.42          & 39.06          & 43.79         & 36.16        & 38.37            & 32.45            \\
			\bottomrule       
		\end{tabular}
	}
\end{table}

\begin{figure}[h]
	\centering
	\includegraphics[width=\textwidth]{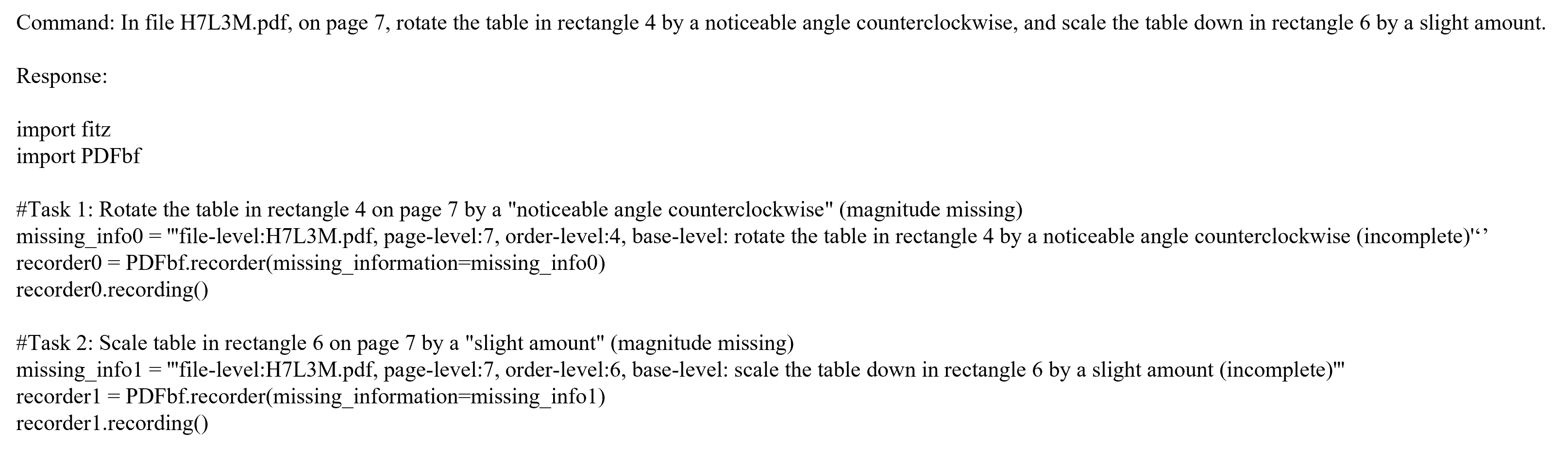}
	\caption{An example of incorrectly recoding vague instruction.}
	\label{fig7}
\end{figure}

As mentioned in the main body, there are two situations that require LLMs critical reasoning to give a correct response, which are incomplete instructions and vague details. For most LLMs, switching from specifying values precisely to vaguely decreases the accuracy of all subtasks (Table \ref{tab7}). The most obvious difficulty introduced is the need to assume arguments in a reasonable way. The most common error is simply filling in the description text instead of assuming a reasonable value. For example, when the instruction asks for a “general font color”, the LLMs will define a variable as follows: “\textit{fontcolor=“general color”}”. Besides, OpenAI o1 is the most sensitive model to vague instructions. Considerable decrease exists not only in argument defining, but also in function calling and tool selection. This is because some values vaguely described are wrongly recognized as missing necessary information. Some vague instructions are incorrectly treated as error instructions. Here is an example response generated by OpenAI o1 shown in Figure \ref{fig7}.

Consistent with the analysis in the main body, there is a very significant decrease in the performance of the models on the tasks set with error instructions in almost all subtasks, except OpenAl o1 (Table \ref{tab8}). However, the reason for OpenAI o1's stable comprehensive score is not a stable accuracy for all subtasks, but there is a significant decrease in plan execution and a small increase in the other subtasks. This means that, although OpenAI o1 can apply the correct response pattern to erroneous instructions, it remains or even becomes more challenging for o1 to fully align with the instructions in detail. This results in more imperfect modifications.

\begin{table}[h]
	\caption{Performance of agents for complete instructions (Com) and incomplete instructions (ERR).}
	\label{tab8}
	\resizebox{\textwidth}{!}{
		\begin{tabular}{lllllllllllll}
			\toprule
			& \multicolumn{2}{c}{OpenAI o1} & \multicolumn{2}{c}{ChatGPT-4o-2024-08-06} & \multicolumn{2}{c}{Claude3.5-sonnet} & \multicolumn{2}{c}{Deepseek-v3-685B} & \multicolumn{2}{c}{Qwen2.5-72B-Instruct} & \multicolumn{2}{c}{Llama3-70B-Instruct} \\ 
			\cmidrule{2-13} 
			& Com           & ERR           & Com            & ERR           & Com               & ERR              & Com            & ERR            & Com           & ERR          & Com              & ERR              \\
			\midrule
			Argument defining       & 60.04         & 63.86         & 68.66          & 46.64         & 71.43             & 51.33            & 72.78          & 38.68          & 75.72         & 24.19        & 74.09            & 15.37            \\
			Variable transferring   & 71.41         & 73.52         & 77.09          & 57.18         & 74.56             & 59.16            & 82.40          & 44.13          & 82.62         & 28.75        & 78.91            & 8.64             \\
			Function calling        & 72.98         & 78.54         & 78.41          & 60.85         & 76.74             & 65.69            & 82.95          & 52.64          & 85.94         & 42.68        & 84.24            & 26.63            \\
			(Single) Tool selection & 66.69         & 75.59         & 72.27          & 47.95         & 78.23             & 54.41            & 78.61          & 34.97          & 80.52         & 18.94        & 78.78            & 14.41            \\
			Multi-tool selection & 69.97         & 76.15         & 78.64          & 71.79         & 63.69             & 75.13            & 79.15          & 67.69          & 84.42         & 61.28        & 82.91            & 4.87             \\
			Plan execution          & 59.66         & 35.87         & 60.51          & 30.11         & 56.90             & 29.72            & 61.18          & 26.40          & 59.36         & 21.97        & 57.72            & 17.31            \\ 
			\bottomrule       
		\end{tabular}
	}
\end{table}

\section{Default Prompts}

All default prompts have four components: task background, standard pipelines(task plan), available functions, and examples:

Task background: This section describes the role of LLM and provides prior knowledge and implicit policy. It includes definitions of terminology, terms defined by the tool designer or user, and other necessary information.

Standard pipeline: This section is important in improving the agent's performance, especially in integrating and utilizing functions. It describes a graph containing multi-mode operations to implement the revision of an object, such as branch shift, step skipping, single tool selection, and multi-tool selection from a sub-library. Instead of working as a plan for a particular task, it works for all possible situations. It supports parallel implementation for instructions with multiple objects, or loop execution for an object with long operations.

Available functions: Detailed information about the available functions is provided in this section, such as function names, calling statements, application scenarios, and the function's essential arguments, optional arguments, input data types, and output results. It is also possible to provide complementary implicit policies if necessary.

Examples: After experimentation, it was found that it is still a challenge for LLMs to follow the pipeline and call functions to complete the instruction in a zero-shot manner. Thus, one or several examples are recommended to help illustrate how pipelines are implemented. 

Here are the default prompts:

\subsection{Adding Text}

\begin{figure}[h]
	\includegraphics[width=\textwidth, viewport=39 1740 1820 4220, clip]{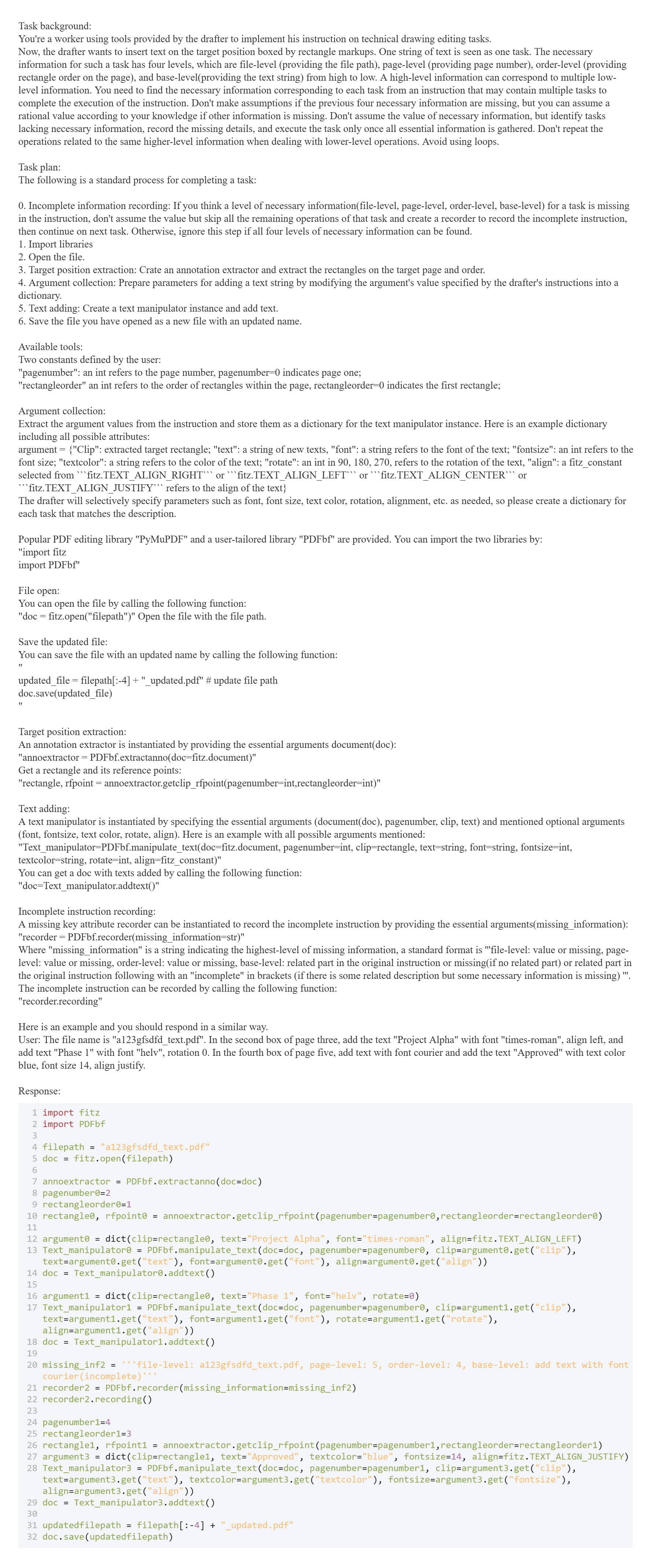}
    \vspace{-0.5cm}
\end{figure}
\clearpage
\begin{figure}[t]
	\includegraphics[width=\textwidth, viewport=39 0 1820 1740, clip]{Adding_text}
    \vspace{-0.5cm}
\end{figure}

\subsection{Revising Text}
\begin{figure}[h]
    \includegraphics[width=\textwidth, viewport=39 4300 1820 5000, clip]{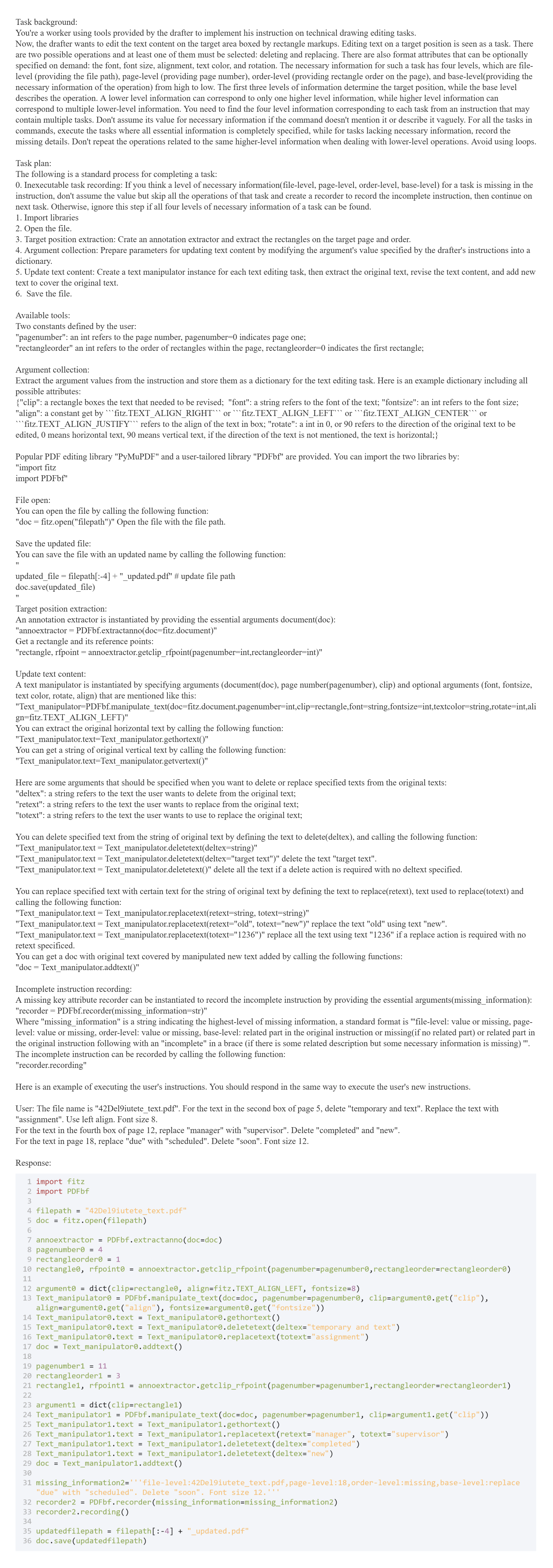}
    \vspace{-0.5cm}
\end{figure}
\clearpage
\begin{figure}[t]
    \includegraphics[width=\textwidth, viewport=39 1600 1820 4300, clip]{Revising_text}
    \vspace{-0.5cm}
\end{figure}
\clearpage
\begin{figure}[t]
    \includegraphics[width=\textwidth, viewport=39 0 1820 1580, clip]{Revising_text}
    \vspace{-0.5cm}
\end{figure}
\subsection{Mapping Text}
\begin{figure}[h]
    \includegraphics[width=\textwidth, viewport=39 2880 1820 3800, clip]{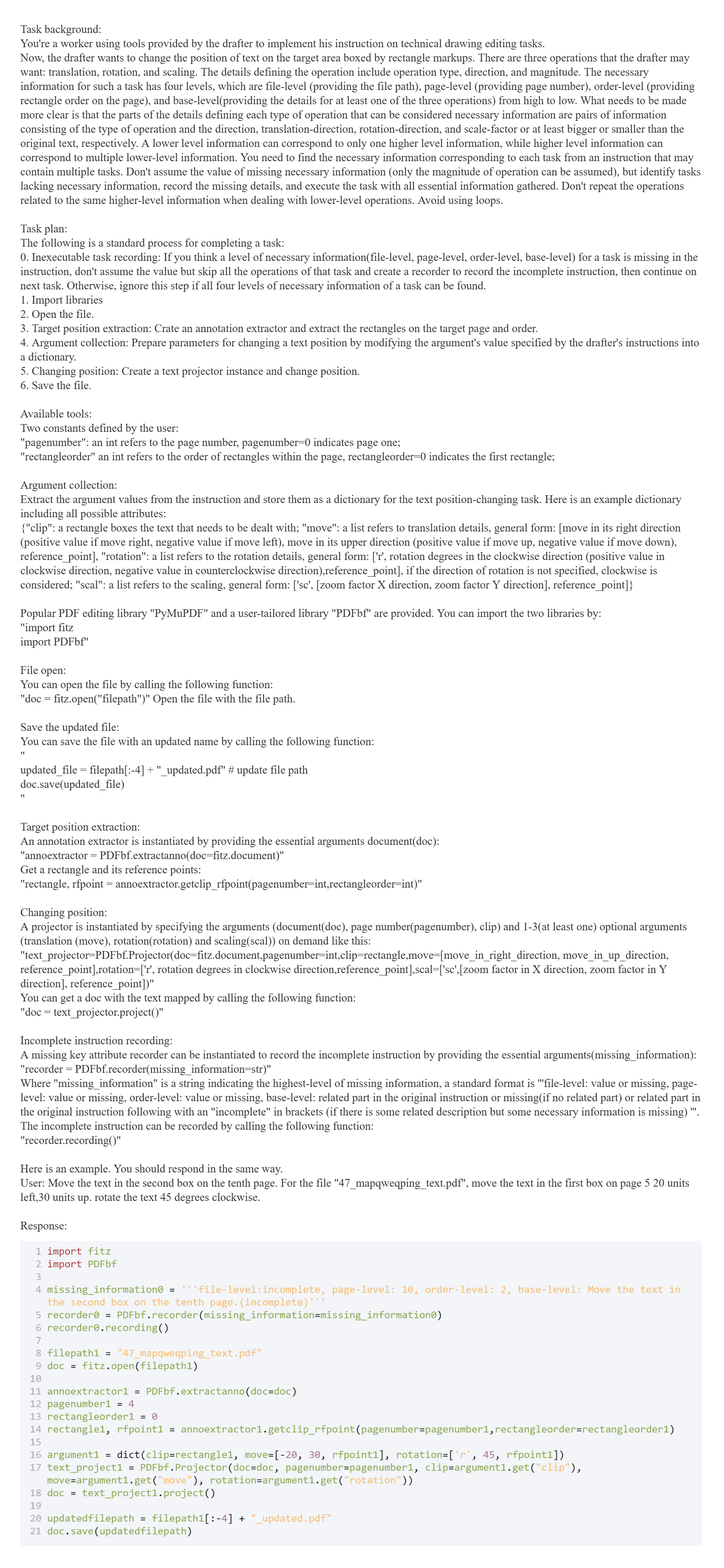}
    \vspace{-0.5cm}
\end{figure}

\clearpage
\begin{figure}[t]
    \includegraphics[width=\textwidth, viewport=39 10 1820 2880, clip]{Mapping_text}
    \vspace{-0.5cm}
\end{figure}
\clearpage
\subsection{Updating Text Format}
\begin{figure}[h]
    \includegraphics[width=\textwidth, viewport=39 1600 1820 4160, clip]{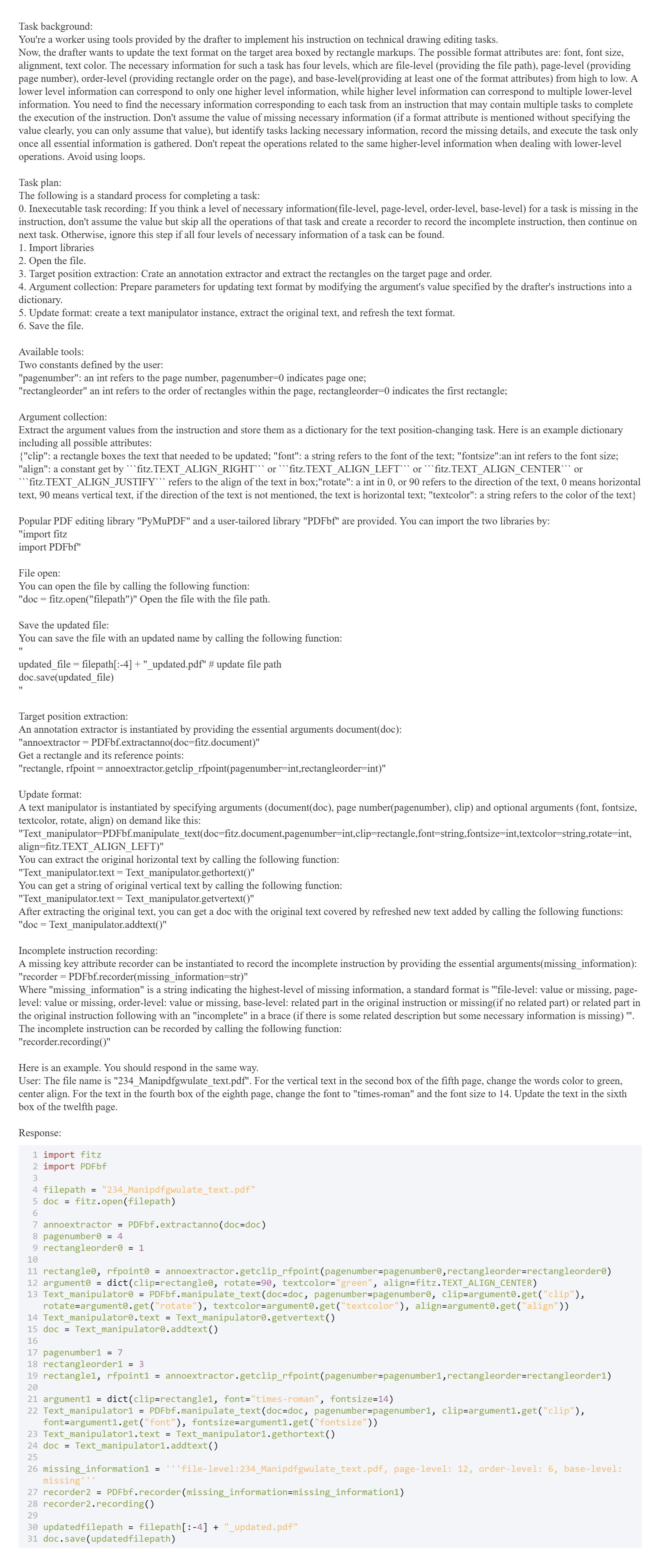}
    \vspace{-0.5cm}
\end{figure}
\clearpage
\begin{figure}[t]
    \includegraphics[width=\textwidth, viewport=39 10 1820 1600, clip]{Refreshing_text}
    \vspace{-0.5cm}
\end{figure}

\subsection{Adding Table}
\begin{figure}[h]
    \includegraphics[width=\textwidth, viewport=39 3700 1820 4600, clip]{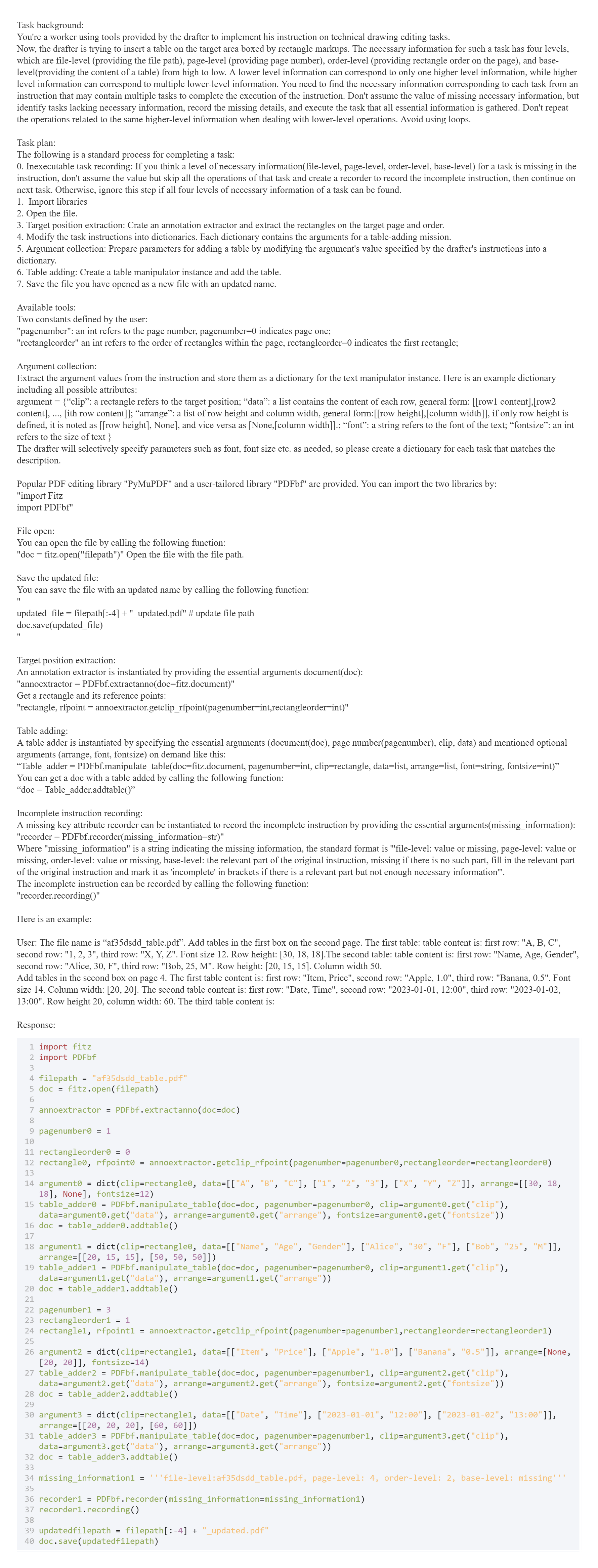}
    \vspace{-0.5cm}
\end{figure}

\clearpage
\begin{figure}[t]
    \includegraphics[width=\textwidth, viewport=39 1000 1820 3700, clip]{Adding_table}
    \vspace{-0.5cm}
\end{figure}
\clearpage
\begin{figure}[t]
    \includegraphics[width=\textwidth, viewport=39 10 1820 980, clip]{Adding_table}
    \vspace{-0.5cm}
\end{figure}
\subsection{Revising Table}
\begin{figure}[h]
    \includegraphics[width=\textwidth, viewport=39 4800 1820 6300, clip]{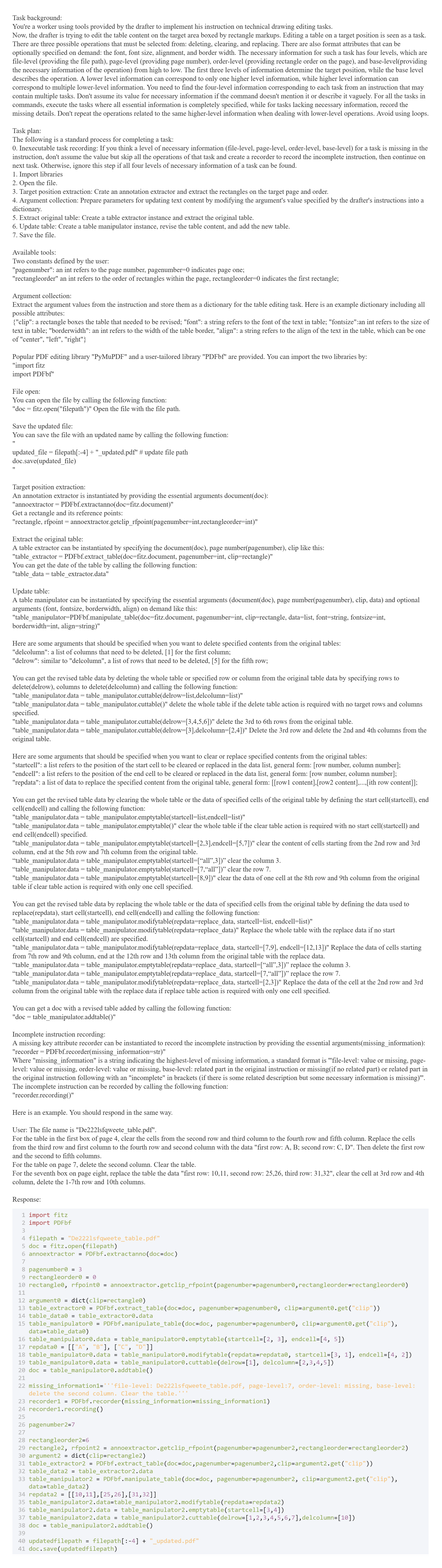}
    \vspace{-0.5cm}
\end{figure}

\clearpage
\begin{figure}[t]
    \includegraphics[width=\textwidth, viewport=39 1990 1820 4750, clip]{Revising_table}
    \vspace{-0.5cm}
\end{figure}

\clearpage
\begin{figure}[t]
    \includegraphics[width=\textwidth, viewport=39 10 1820 1990, clip]{Revising_table}
    \vspace{-0.5cm}
\end{figure}
\subsection{Mapping Table}
\begin{figure}[h]
    \includegraphics[width=\textwidth, viewport=39 3700 1820 4200, clip]{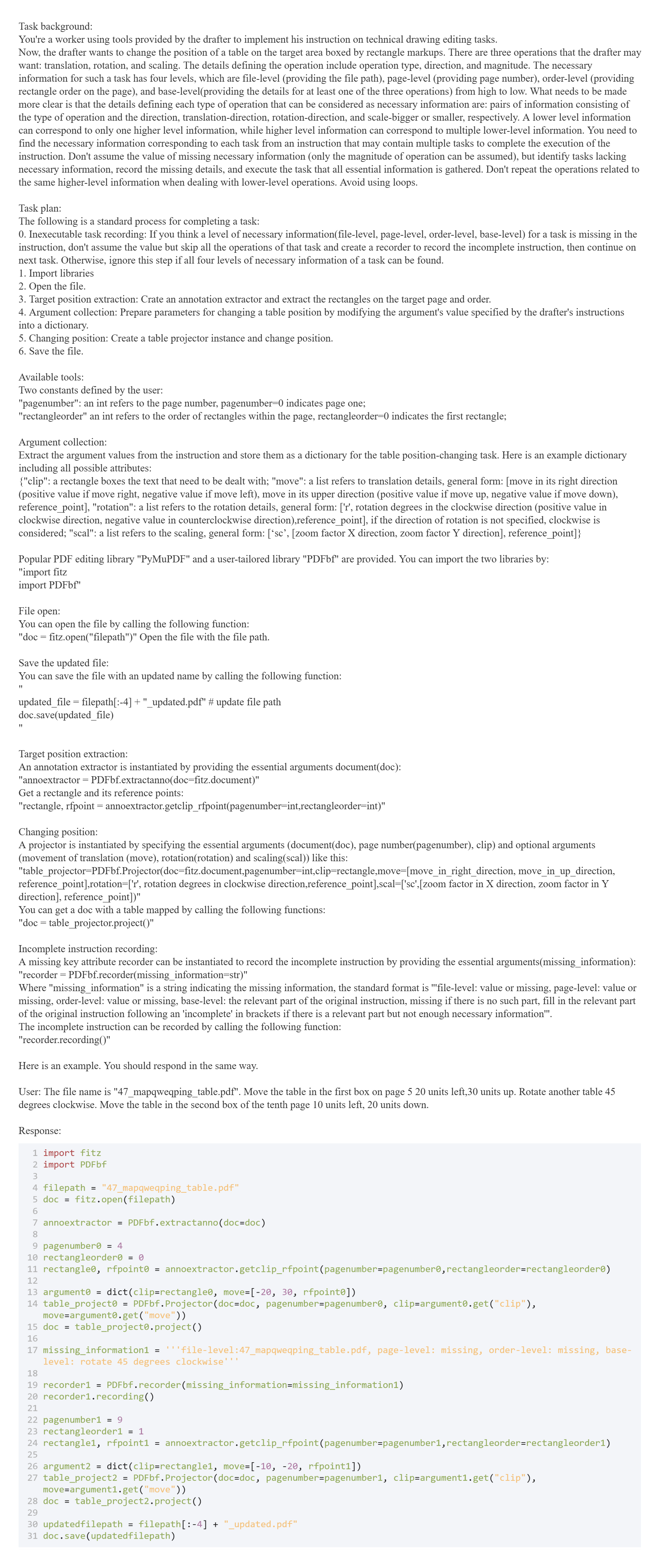}
    \vspace{-0.5cm}
\end{figure}

\clearpage
\begin{figure}[t]
    \includegraphics[width=\textwidth, viewport=39 910 1820 3680, clip]{Mapping_table}
    \vspace{-0.5cm}
\end{figure}
\clearpage
\begin{figure}[t]
    \includegraphics[width=\textwidth, viewport=39 10 1820 890, clip]{Mapping_table}
    \vspace{-0.5cm}
\end{figure}
\subsection{Updating Table Format}
\begin{figure}[h]
    \includegraphics[width=\textwidth, viewport=39 2300 1820 3920, clip]{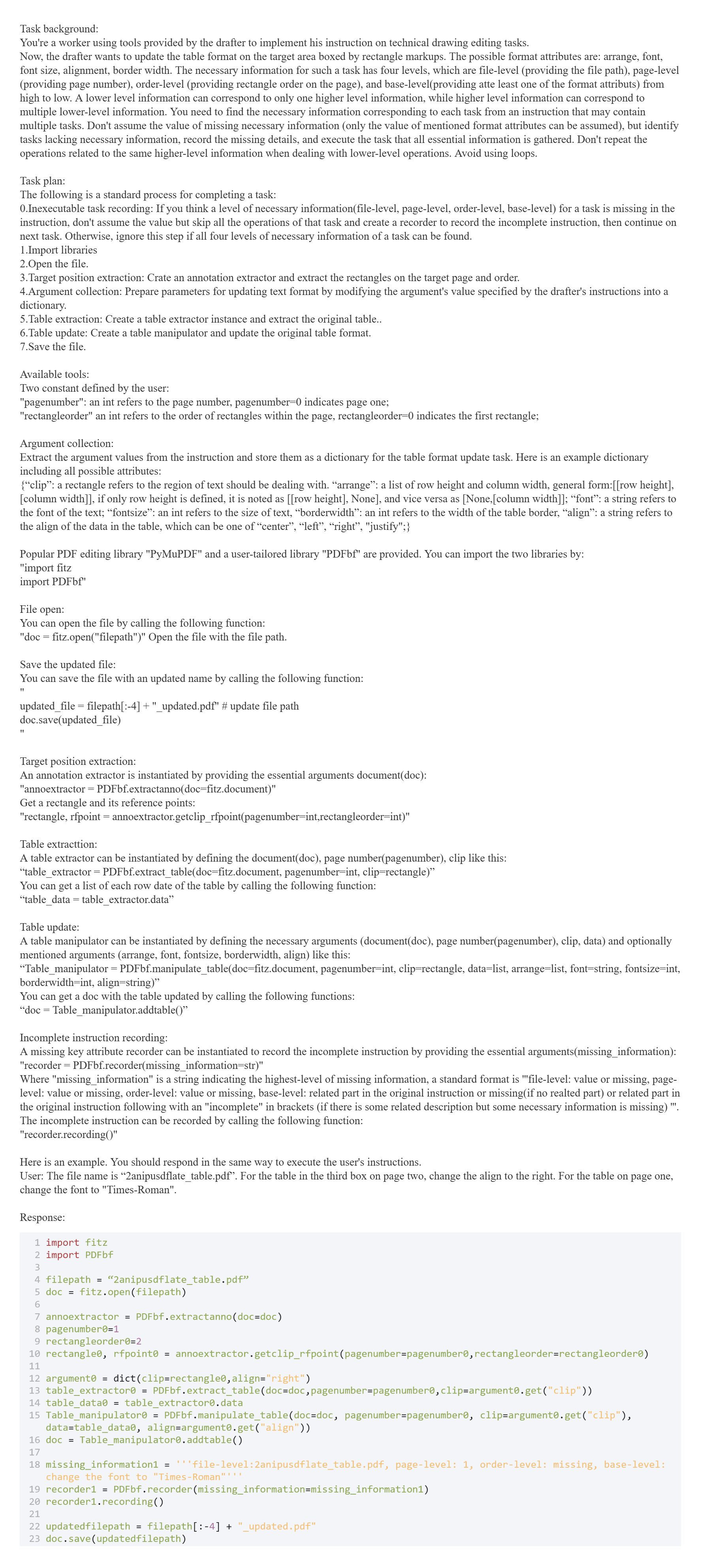}
    \vspace{-0.5cm}
\end{figure}
\clearpage
\begin{figure}[t]
    \includegraphics[width=\textwidth, viewport=39 10 1820 2300, clip]{Refreshing_table}
    \vspace{-0.5cm}
\end{figure}
\subsection{Adding Vectors}
\begin{figure}[h]
    \includegraphics[width=\textwidth, viewport=39 4575 1820 4793, clip]{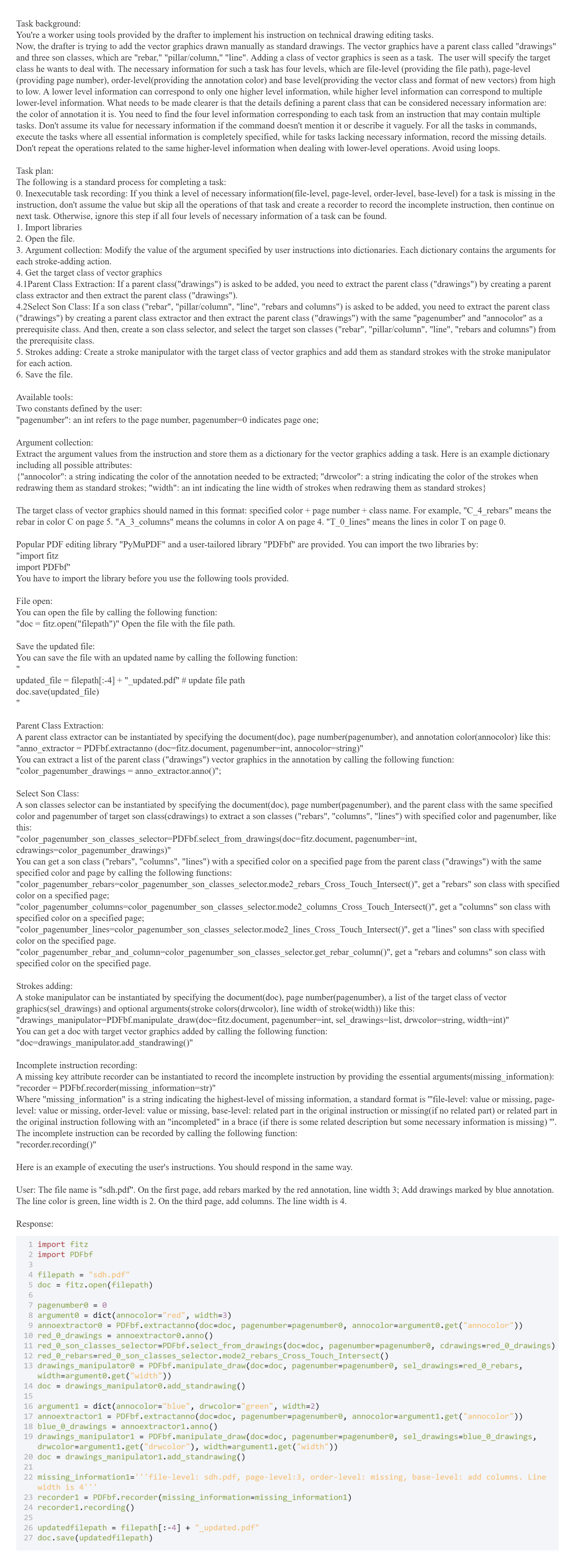}
    \vspace{-0.5cm}
\end{figure}
\clearpage
\begin{figure}[h]
    \includegraphics[width=\textwidth, viewport=39 1920 1820 4575, clip]{Adding_vectors}
    \vspace{-0.5cm}
\end{figure}
\clearpage
\begin{figure}[t]
    \includegraphics[width=\textwidth, viewport=39 10 1820 1920, clip]{Adding_vectors}
    \vspace{-0.5cm}
\end{figure}
\subsection{Deleting Vectors}
\begin{figure}[h]
    \includegraphics[width=\textwidth, viewport=39 5200 1820 5700, clip]{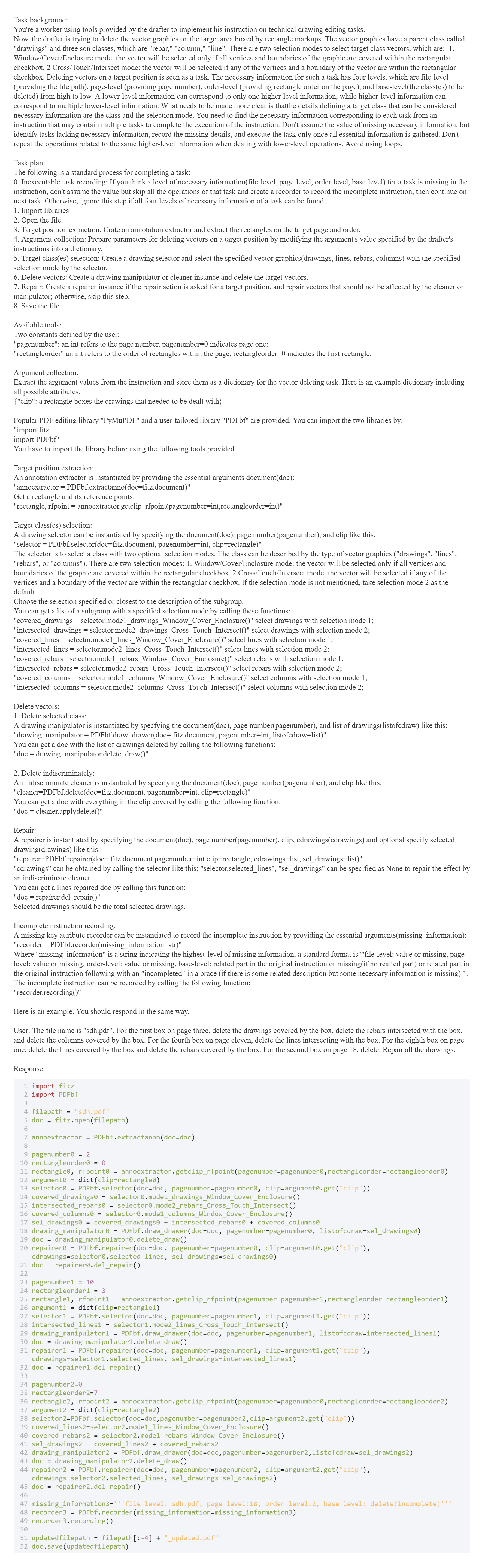}
    \vspace{-0.5cm}
\end{figure}

\clearpage
\begin{figure}[h]
    \includegraphics[width=\textwidth, viewport=39 2400 1820 5180, clip]{Delete_vectors}
    \vspace{-0.5cm}
\end{figure}

\clearpage
\begin{figure}[t]
    \includegraphics[width=\textwidth, viewport=39 10 1820 2400, clip]{Delete_vectors}
    \vspace{-0.5cm}
\end{figure}
\subsection{Mapping Vectors}
\begin{figure}[h]
    \includegraphics[width=\textwidth, viewport=39 5930 1820 6030, clip]{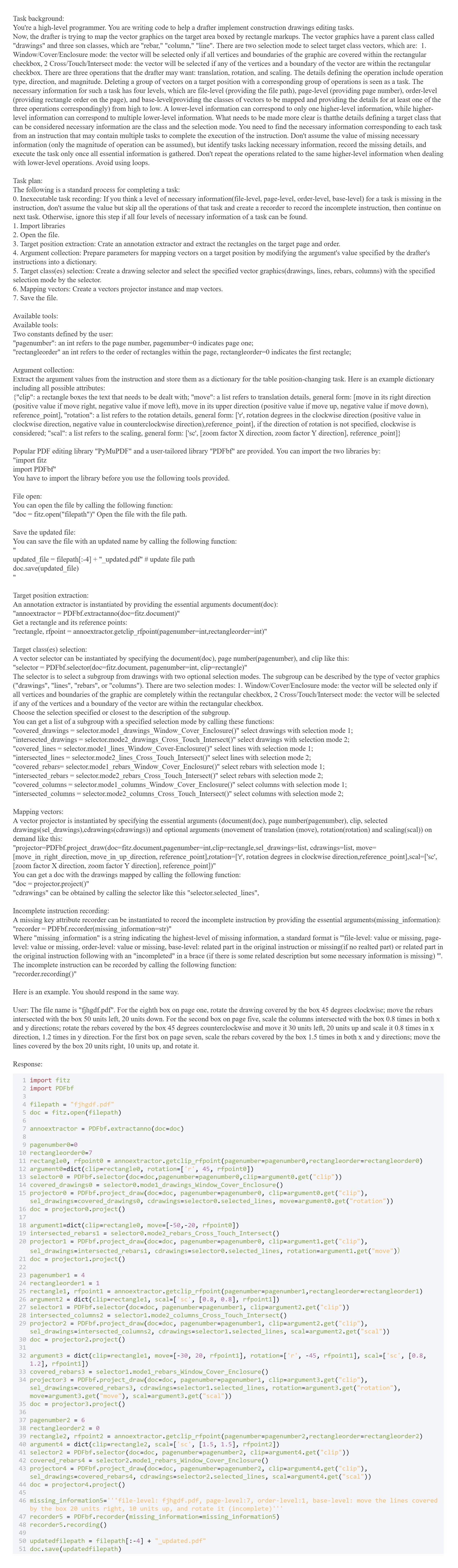}
    \vspace{-0.5cm}
\end{figure}

\clearpage
\begin{figure}[h]
    \includegraphics[width=\textwidth, viewport=39 3090 1820 5930, clip]{Mapping_vectors}
    \vspace{-0.5cm}
\end{figure}

\clearpage
\begin{figure}[h]
    \includegraphics[width=\textwidth, viewport=39 310 1820 3090, clip]{Mapping_vectors}
    \vspace{-0.5cm}
\end{figure}

\clearpage
\begin{figure}[t]
    \includegraphics[width=\textwidth, viewport=39 10 1820 280, clip]{Mapping_vectors}
    \vspace{-0.5cm}
\end{figure}

\subsection{Updating Vector Format}
\begin{figure}[h]
    \includegraphics[width=\textwidth, viewport=39 3700 1820 5930, clip]{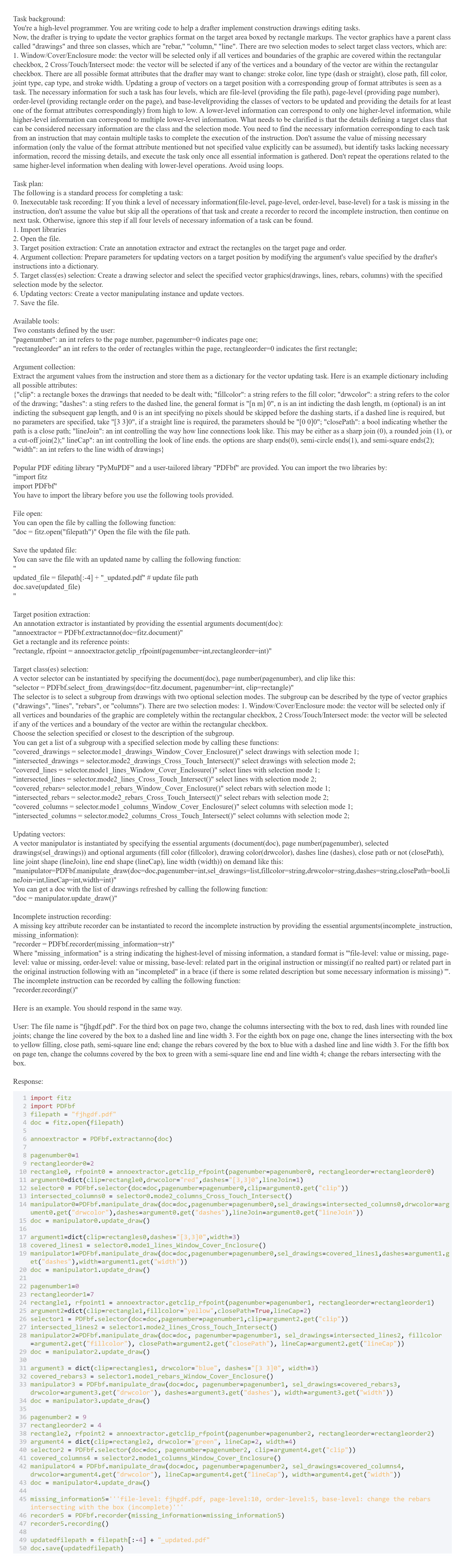}
    \vspace{-0.5cm}
\end{figure}

\clearpage
\begin{figure}[h]
    \includegraphics[width=\textwidth, viewport=39 930 1820 3670, clip]{Refreshing_vectors}
    \vspace{-0.5cm}
\end{figure}

\clearpage
\begin{figure}[t]
    \includegraphics[width=\textwidth, viewport=39 10 1820 930, clip]{Refreshing_vectors}
    \vspace{-0.5cm}
\end{figure}
Here are all the default prompts. We encourage users to develop their own prompts to achieve higher scores.


\begin{thebibliography}{39}
\providecommand{\natexlab}[1]{#1}
\providecommand{\url}[1]{\texttt{#1}}
\expandafter\ifx\csname urlstyle\endcsname\relax
  \providecommand{\doi}[1]{doi: #1}\else
  \providecommand{\doi}{doi: \begingroup \urlstyle{rm}\Url}\fi

\bibitem[Ahn et~al.(2022)Ahn, Brohan, Brown, Chebotar, Cortes, David, Finn, Fu, Gopalakrishnan, and Hausman]{RN7}
Michael Ahn, Anthony Brohan, Noah Brown, Yevgen Chebotar, Omar Cortes, Byron David, Chelsea Finn, Chuyuan Fu, Keerthana Gopalakrishnan, and Karol Hausman.
\newblock Do as i can, not as i say: Grounding language in robotic affordances.
\newblock \emph{arXiv preprint arXiv:2204.01691}, 2022.

\bibitem[Anthrop(2024)]{RN66}
Anthrop.
\newblock Claude3.5-sonnet, 2024.
\newblock URL \url{https://www.llama.com/}.

\bibitem[Artifex(2025)]{RN42}
Artifex.
\newblock Pymupdf, 2025.
\newblock URL \url{https://pypi.org/project/PyMuPDF/}.

\bibitem[Besta et~al.(2023)Besta, Blach, Kubicek, Gerstenberger, Podstawski, Gianinazzi, Gajda, Lehmann, Niewiadomski, and Nyczyk]{RN1}
Maciej Besta, Nils Blach, Ales Kubicek, Robert Gerstenberger, Michal Podstawski, Lukas Gianinazzi, Joanna Gajda, Tomasz Lehmann, Hubert Niewiadomski, and Piotr Nyczyk.
\newblock Graph of thoughts: Solving elaborate problems with large language models.
\newblock In \emph{Proceedings of the AAAI Conference on Artificial Intelligence}, volume~38, pages 17682--17690, 2023.
\newblock ISBN 2374-3468.

\bibitem[Deepseek(2024)]{RN67}
Deepseek.
\newblock Deepseek v3, 2024.
\newblock URL \url{https://www.deepseek.com/}.

\bibitem[EvolveLab(2025)]{RN23}
EvolveLab.
\newblock Glyph copilot, 2025.
\newblock URL \url{https://www.evolvelab.io/glyph}.

\bibitem[Hoffstaetter(2025)]{RN45}
S.~Hoffstaetter.
\newblock pytesseract, 2025.
\newblock URL \url{https://pypi.org/project/pytesseract/}.

\bibitem[Kim et~al.(2024)Kim, Baldi, and McAleer]{RN10}
Geunwoo Kim, Pierre Baldi, and Stephen McAleer.
\newblock Language models can solve computer tasks.
\newblock \emph{Advances in Neural Information Processing Systems}, 36, 2024.

\bibitem[Kim et~al.(2023)Kim, Moon, Tabrizi, Lee, Mahoney, Keutzer, and Gholami]{RN12}
Sehoon Kim, Suhong Moon, Ryan Tabrizi, Nicholas Lee, Michael~W Mahoney, Kurt Keutzer, and Amir Gholami.
\newblock An llm compiler for parallel function calling.
\newblock \emph{arXiv preprint arXiv:2312.04511}, 2023.

\bibitem[Li et~al.(2023)Li, Zhao, Yu, Song, Li, Yu, Li, Huang, and Li]{RN59}
Minghao Li, Yingxiu Zhao, Bowen Yu, Feifan Song, Hangyu Li, Haiyang Yu, Zhoujun Li, Fei Huang, and Yongbin Li.
\newblock Api-bank: A comprehensive benchmark for tool-augmented llms.
\newblock \emph{arXiv preprint arXiv:2304.08244}, 2023.

\bibitem[Li et~al.(2022)Li, Choi, Chung, Kushman, Schrittwieser, Leblond, Eccles, Keeling, Gimeno, and Dal~Lago]{RN13}
Yujia Li, David Choi, Junyoung Chung, Nate Kushman, Julian Schrittwieser, Rémi Leblond, Tom Eccles, James Keeling, Felix Gimeno, and Agustin Dal~Lago.
\newblock Competition-level code generation with alphacode.
\newblock \emph{Science}, 378\penalty0 (6624):\penalty0 1092--1097, 2022.
\newblock ISSN 0036-8075.

\bibitem[Liu et~al.(2019)Liu, Liu, Feng, Wu, and Lan]{RN22}
Jiepeng Liu, P~Liu, L~Feng, W~Wu, and H~Lan.
\newblock Automated clash resolution of rebar design in rc joints using multi-agent reinforcement learning and bim.
\newblock In \emph{ISARC. Proceedings of the International Symposium on Automation and Robotics in Construction}, volume~36, pages 921--928. IAARC Publications, 2019.

\bibitem[Ma et~al.(2024)Ma, Zhang, Zhu, Yang, Yang, Jin, Lan, Kong, and He]{RN37}
Chang Ma, Junlei Zhang, Zhihao Zhu, Cheng Yang, Yujiu Yang, Yaohui Jin, Zhenzhong Lan, Lingpeng Kong, and Junxian He.
\newblock Agentboard: An analytical evaluation board of multi-turn llm agents.
\newblock \emph{ArXiv}, abs/2401.13178, 2024.

\bibitem[Masterman et~al.(2024)Masterman, Besen, Sawtell, and Chao]{RN46}
Tula Masterman, Sandi Besen, Mason Sawtell, and Alex Chao.
\newblock The landscape of emerging ai agent architectures for reasoning, planning, and tool calling: A survey.
\newblock \emph{ArXiv}, abs/2404.11584, 2024.

\bibitem[Meta(2024)]{RN65}
Meta.
\newblock Llama3.3, 2024.
\newblock URL \url{https://www.llama.com/}.

\bibitem[Mialon et~al.(2023)Mialon, Fourrier, Swift, Wolf, LeCun, and Scialom]{RN41}
Grégoire Mialon, Clémentine Fourrier, Craig Swift, Thomas Wolf, Yann LeCun, and Thomas Scialom.
\newblock Gaia: a benchmark for general ai assistants.
\newblock \emph{ArXiv}, abs/2311.12983, 2023.

\bibitem[OpenAI(2024)]{RN64}
OpenAI.
\newblock Openai o1, 2024.
\newblock URL \url{https://openai.com/index/learning-to-reason-with-llms/}.

\bibitem[OpenCV-Team(2025)]{RN44}
OpenCV-Team.
\newblock Opencv, 2025.
\newblock URL \url{https://pypi.org/project/opencv-python/}.

\bibitem[Paranjape et~al.(2023)Paranjape, Lundberg, Singh, Hajishirzi, Zettlemoyer, and Ribeiro]{RN53}
Bhargavi Paranjape, Scott Lundberg, Sameer Singh, Hannaneh Hajishirzi, Luke Zettlemoyer, and Marco~Tulio Ribeiro.
\newblock Art: Automatic multi-step reasoning and tool-use for large language models.
\newblock \emph{arXiv preprint arXiv:2303.09014}, 2023.

\bibitem[Qin et~al.(2023)Qin, Liang, Ye, Zhu, Yan, Lu, Lin, Cong, Tang, and Qian]{RN30}
Yujia Qin, Shihao Liang, Yining Ye, Kunlun Zhu, Lan Yan, Yaxi Lu, Yankai Lin, Xin Cong, Xiangru Tang, and Bill Qian.
\newblock Toolllm: Facilitating large language models to master 16000+ real-world apis.
\newblock \emph{arXiv preprint arXiv:2307.16789}, 2023.

\bibitem[Qwen-Team(2024)]{RN68}
Qwen-Team.
\newblock Qwen2.5, 2024.
\newblock URL \url{https://qwen2.org/qwen2-5/}.

\bibitem[ReportLab(2025)]{RN43}
ReportLab.
\newblock Reportlab, 2025.
\newblock URL \url{https://docs.reportlab.com/}.

\bibitem[Shi et~al.(2024)Shi, Gao, Chen, Feng, Yan, Shi, Yin, Ren, Verberne, and Ren]{RN56}
Zhengliang Shi, Shen Gao, Xiuyi Chen, Yue Feng, Lingyong Yan, Haibo Shi, Dawei Yin, Pengjie Ren, Suzan Verberne, and Zhaochun Ren.
\newblock Learning to use tools via cooperative and interactive agents.
\newblock \emph{arXiv preprint arXiv:2403.03031}, 2024.

\bibitem[Song et~al.(2023)Song, Xiong, Zhu, Wu, Qian, Song, Huang, Li, Wang, and Yao]{RN58}
Yifan Song, Weimin Xiong, Dawei Zhu, Wenhao Wu, Han Qian, Mingbo Song, Hailiang Huang, Cheng Li, Ke~Wang, and Rong Yao.
\newblock Restgpt: Connecting large language models with real-world restful apis.
\newblock \emph{arXiv preprint arXiv:2306.06624}, 2023.

\bibitem[Srinivasan et~al.(2023)Srinivasan, Dong, Zhu, Yu, Mosk-Aoyama, Keutzer, Jiao, and Zhang]{RN11}
Venkat~Krishna Srinivasan, Zhen Dong, Banghua Zhu, Brian Yu, Damon Mosk-Aoyama, Kurt Keutzer, Jiantao Jiao, and Jian Zhang.
\newblock Nexusraven: a commercially-permissive language model for function calling.
\newblock In \emph{NeurIPS 2023 Foundation Models for Decision Making Workshop}, 2023.

\bibitem[Srivastava et~al.(2022)Srivastava, Rastogi, Rao, Shoeb, Abid, Fisch, Brown, Santoro, Gupta, and Garriga-Alonso]{RN34}
Aarohi Srivastava, Abhinav Rastogi, Abhishek Rao, Abu Awal~Md Shoeb, Abubakar Abid, Adam Fisch, Adam~R Brown, Adam Santoro, Aditya Gupta, and Adrià Garriga-Alonso.
\newblock Beyond the imitation game: Quantifying and extrapolating the capabilities of language models.
\newblock \emph{arXiv preprint arXiv:2206.04615}, 2022.

\bibitem[Wang et~al.(2023)Wang, Wang, Liu, Chen, Yuan, Peng, and Ji]{RN39}
Xingyao Wang, Zihan Wang, Jiateng Liu, Yangyi Chen, Lifan Yuan, Hao Peng, and Heng Ji.
\newblock Mint: Evaluating llms in multi-turn interaction with tools and language feedback.
\newblock \emph{ArXiv}, abs/2309.10691, 2023.

\bibitem[Wang et~al.(2024)Wang, Chen, Yuan, Zhang, Li, Peng, and Ji]{RN73}
Xingyao Wang, Yangyi Chen, Lifan Yuan, Yizhe Zhang, Yunzhu Li, Hao Peng, and Heng Ji.
\newblock Executable code actions elicit better llm agents.
\newblock In \emph{Forty-first International Conference on Machine Learning}, 2024.

\bibitem[Wei et~al.(2022)Wei, Wang, Schuurmans, Bosma, Xia, Chi, Le, and Zhou]{RN5}
Jason Wei, Xuezhi Wang, Dale Schuurmans, Maarten Bosma, Fei Xia, Ed~Chi, Quoc~V Le, and Denny Zhou.
\newblock Chain-of-thought prompting elicits reasoning in large language models.
\newblock \emph{Advances in neural information processing systems}, 35:\penalty0 24824--24837, 2022.

\bibitem[Wu et~al.(2023{\natexlab{a}})Wu, Min, Bisk, Salakhutdinov, Azaria, Li, Mitchell, and Prabhumoye]{RN8}
Yue Wu, So~Yeon Min, Yonatan Bisk, Ruslan Salakhutdinov, Amos Azaria, Yuanzhi Li, Tom Mitchell, and Shrimai Prabhumoye.
\newblock Plan, eliminate, and track--language models are good teachers for embodied agents.
\newblock \emph{arXiv preprint arXiv:2305.02412}, 2023{\natexlab{a}}.

\bibitem[Wu et~al.(2023{\natexlab{b}})Wu, Tang, Mitchell, and Li]{RN33}
Yue Wu, Xuan Tang, Tom~M Mitchell, and Yuanzhi Li.
\newblock Smartplay: A benchmark for llms as intelligent agents.
\newblock \emph{arXiv preprint arXiv:2310.01557}, 2023{\natexlab{b}}.

\bibitem[Xu et~al.(2023)Xu, Hong, Li, Hu, Chen, and Zhang]{RN57}
Qiantong Xu, Fenglu Hong, Bo~Li, Changran Hu, Zhengyu Chen, and Jian Zhang.
\newblock On the tool manipulation capability of open-source large language models.
\newblock \emph{arXiv preprint arXiv:2305.16504}, 2023.

\bibitem[Xue et~al.(2024)Xue, Chen, Zhou, Dai, Chu, and Mei]{RN36}
Siqiao Xue, Tingting Chen, Fan Zhou, Qingyang Dai, Zhixuan Chu, and Hongyuan Mei.
\newblock Famma: A benchmark for financial domain multilingual multimodal question answering.
\newblock \emph{arXiv preprint arXiv:2410.04526}, 2024.

\bibitem[Yan et~al.(2024)Yan, Mao, Ji, Zhang, Patil, Stoica, and Gonzalez]{RN62}
Fanjia Yan, Huanzhi Mao, Charlie Cheng-Jie Ji, Tianjun Zhang, Shishir~G. Patil, Ion Stoica, and Joseph~E. Gonzalez.
\newblock Berkeley function calling leaderboard, 2024.
\newblock URL \url{https://gorilla.cs.berkeley.edu/blogs/8_berkeley_function_calling_leaderboard.html}.

\bibitem[Yao et~al.(2022)Yao, Zhao, Yu, Du, Shafran, Narasimhan, and Cao]{RN15}
Shunyu Yao, Jeffrey Zhao, Dian Yu, Nan Du, Izhak Shafran, Karthik Narasimhan, and Yuan Cao.
\newblock React: Synergizing reasoning and acting in language models.
\newblock \emph{arXiv preprint arXiv:2210.03629}, 2022.

\bibitem[Yao et~al.(2024)Yao, Yu, Zhao, Shafran, Griffiths, Cao, and Narasimhan]{RN6}
Shunyu Yao, Dian Yu, Jeffrey Zhao, Izhak Shafran, Tom Griffiths, Yuan Cao, and Karthik Narasimhan.
\newblock Tree of thoughts: Deliberate problem solving with large language models.
\newblock \emph{Advances in Neural Information Processing Systems}, 36, 2024.

\bibitem[Yin et~al.(2024{\natexlab{a}})Yin, Bai, Ma, Nan, Sun, Xu, Ma, Lu, Kong, and Zhang]{RN31}
Guoli Yin, Haoping Bai, Shuang Ma, Feng Nan, Yanchao Sun, Zhaoyang Xu, Shen Ma, Jiarui Lu, Xiang Kong, and Aonan Zhang.
\newblock Mmau: A holistic benchmark of agent capabilities across diverse domains.
\newblock \emph{arXiv preprint arXiv:2407.18961}, 2024{\natexlab{a}}.

\bibitem[Yin et~al.(2024{\natexlab{b}})Yin, Bai, Ma, Nan, Sun, Xu, Ma, Lu, Kong, and Zhang]{RN72}
Guoli Yin, Haoping Bai, Shuang Ma, Feng Nan, Yanchao Sun, Zhaoyang Xu, Shen Ma, Jiarui Lu, Xiang Kong, and Aonan Zhang.
\newblock Mmau: A holistic benchmark of agent capabilities across diverse domains.
\newblock \emph{arXiv preprint arXiv:2407.18961}, 2024{\natexlab{b}}.

\bibitem[Zhuo et~al.(2024)Zhuo, Vu, Chim, Hu, Yu, Widyasari, Yusuf, Zhan, He, and Paul]{RN35}
Terry~Yue Zhuo, Minh~Chien Vu, Jenny Chim, Han Hu, Wenhao Yu, Ratnadira Widyasari, Imam Nur~Bani Yusuf, Haolan Zhan, Junda He, and Indraneil Paul.
\newblock Bigcodebench: Benchmarking code generation with diverse function calls and complex instructions.
\newblock \emph{arXiv preprint arXiv:2406.15877}, 2024.

\end{thebibliography}
\end{document}